\documentclass[manuscript,screen,nonacm]{acmart}

\AtBeginDocument{%
  }

\usepackage{amsmath}
\usepackage{algorithm}
\usepackage{algpseudocode}
\usepackage{tikz}
\usetikzlibrary{fit, calc, shapes.arrows, shapes.geometric}
\usepackage{graphicx}
\usepackage{subcaption}
\usepackage{csquotes}
\usepackage{wrapfig}
\usepackage{import}
\usepackage{multirow}

\newcommand{\Dtrain}{\mathcal{D}_\text{train}}
\newcommand{\Dtest}{\mathcal{D}_\text{test}}
\newcommand{\vx}{\mathbf{x}}
\newcommand{\vX}{\mathbf{X}}
\newcommand{\vU}{\mathbf{U}}
\newcommand{\xtrain}{\mathcal{X}_\text{train}}
\newcommand{\xtest}{\mathcal{X}_\text{test}}

\newcommand{\ytest}{\mathcal{Y}_\text{test}}
\newcommand{\vpi}{\hat{\boldsymbol{\pi}}}
\newcommand{\pitrain}{\vpi_\text{train}}
\newcommand{\pitest}{\vpi_\text{test}}
\newcommand{\vdelta}{\hat{\boldsymbol{\delta}}}
\algrenewcommand\algorithmicrequire{\textbf{Input}}
\algrenewcommand\algorithmicensure{\textbf{Output}}

\begin{document}

\title{Overcoming Fairness Trade-offs via Pre-processing: A Causal Perspective}

\author{Charlotte Leininger}
\email{C.Leininger@campus.lmu.de}
\orcid{0009-0004-7492-2777}
\affiliation{%
  \institution{LMU Munich}
  \city{Munich}
  \country{Germany}}

\author{Simon Rittel}
\email{simon.rittel@stat.uni-muenchen.de}
\affiliation{%
  \institution{LMU Munich}
  \city{Munich}
  \country{Germany}
}
\affiliation{%
  \institution{Munich Center for Machine Learning (MCML)}
  \city{Munich}
  \country{Germany}}

\author{Ludwig Bothmann}
\email{ludwig.bothmann@lmu.de}
\orcid{0000-0002-1471-6582}
\affiliation{%
  \institution{LMU Munich}
  \city{Munich}
  \country{Germany}
}
\affiliation{%
  \institution{Munich Center for Machine Learning (MCML)}
  \city{Munich}
  \country{Germany}}

\begin{abstract}
Training machine learning models for fair decisions faces two key challenges: The \emph{fairness-accuracy trade-off} results from enforcing fairness which weakens its predictive performance in contrast to an unconstrained model. The incompatibility of different fairness metrics poses 
another trade-off -- also known as the \emph{impossibility theorem}.
Recent work
identifies the bias within the observed data as a possible root cause and shows that fairness and predictive performance are in fact in accord when predictive performance is measured on unbiased data. We offer a causal explanation for these findings using the framework of the 
FiND (fictitious and normatively desired) world, a ``fair'' world, where protected attributes have no causal effects on the target variable.
We show theoretically that (i) classical fairness metrics deemed to be incompatible are naturally satisfied in the FiND world, while (ii) fairness aligns with high predictive performance. 
We extend our analysis by suggesting how one can benefit from these theoretical insights in practice, using causal pre-processing methods that approximate the FiND world.
Additionally, we propose a method for evaluating the approximation of the FiND world via pre-processing in practical use cases where we do not have access to the FiND world.
In simulations and empirical studies, we demonstrate that these pre-processing methods are successful in approximating the FiND world 
and resolve both trade-offs. 
Our results provide actionable solutions for practitioners to achieve fairness and high predictive performance simultaneously.

\end{abstract}

\ccsdesc[500]{Social and professional topics}
\ccsdesc[500]{Social and professional topics~Computing / technology policy}
\ccsdesc[300]{Applied computing~Law, social and behavioral sciences}
\ccsdesc[100]{Computing methodologies~Machine learning}

\keywords{Fairness-accuracy trade-off, impossibility theorem, causal fairness, bias mitigation, pre-processing}

\maketitle

\section{Introduction}
\label{sec:introduction}
The use of automated decision-making (ADM) systems has become increasingly popular in a variety of fields that were previously solely controlled by humans, including sensitive areas such as loan applications~\cite{kozodoi_credit_20}, hiring choices~\cite{faliagka_rec_2012}, and the criminal justice system~\cite{angwin2016machine}. Such systems have been shown to suffer from bias with respect to certain protected attributes (PAs) -- such as gender or race -- which is in conflict with a variety of anti-discrimination laws, such as the US Civil Rights Act of 1964 or the Charter of Fundamental Rights of the European Union. 
In response, this concern has led to a growing body of literature on fairness-aware machine learning (fairML), proposing various methods to measure and achieve fairness for ADM systems based on machine learning (ML) models. 

This was followed by the observation that fulfilling a fairness notion comes with a decrease in accuracy -- the  \emph{fairness-accuracy trade-off} (see, e.g.,~\cite{cooper_emergent_2021, corbett_cost_2017, hardt_eo_2016}).
Concurrently, with the multitude of suggested fairness metrics came the realization that some of them cannot be met at the same time. There appears to be a mathematical trade-off between several fairness metrics, that cannot be satisfied
simultaneously unless in some special cases -- 
also referred to as the \emph{impossibility theorem} \citep{kleinberg_inherent_2017, chouldechova_fair_2017}.
Newer studies have begun to question these seemingly inevitable trade-offs, particularly in relation to the potential bias present in the data~\cite{wick_unlocking_2019, dutta_is_2020, blum_recovering_2020}. When fairness and accuracy are measured on such biased data, the results must be assumed to be similarly exposed to bias. This therefore raises the need to eliminate bias from the data and evaluate these metrics on unbiased data instead. In doing so, it suggests that fairness and accuracy can actually enhance each other rather than conflict. Similarly, by obtaining unbiased 
data that represents a ``fair world'', measuring fairness with respect to different fairness metrics should yield consistent results, rather than conflicting ones.

\subsection{Our Contributions}
\label{sec:contributons}
In this paper, we explain prior findings that the fairness-accuracy trade-off is resolved when predictive performance is evaluated on unbiased data from a causal viewpoint. Therefore, we investigate the trade-offs in a causal framework of a ``fair world'' and utilize the FiND world -- a ``fictitious and normatively desired'' world proposed by \citet{bothmann_what_2024}. They provide a philosophically sound definition of a fair world, where the PAs have no causal effect on the target, neither directly nor indirectly. We examine the theoretical implications of this FiND world and demonstrate:
\begin{itemize}
    \item \textbf{Theoretical Resolution of the Trade-offs:} First, we show that the FiND world ensures fairness at both the individual and the group level, overcoming the trade-off between incompatible fairness metrics.  This is due to the fact that multiple fairness metrics are naturally satisfied in this world and groups' base rates are equal -- which is a special case of the impossibility theorem. 
    Second, it allows us to overcome the fairness-accuracy trade-off by explaining how enforcing a fairness metric on a model trained in the real world leads to improved predictive performance in the FiND world. 
\end{itemize}
Furthermore, we translate this theory into practice. Since we do not have access to the FiND world in real life, we need to approximate it. For this, we compare two causal pre-processing approaches: the \textit{fairadapt}~\cite{plecko_fair_2020} method and the \textit{residual-based warping} method of~\cite{bothmann_causal_2024}. We demonstrate the effectiveness of the pre-processing methods in a simulation study by evaluating these approximations against the true FiND world: 
\begin{itemize} 
    \item \textbf{Resolving the Trade-Offs Practically via Pre-Processing:} We propose a method to evaluate whether the pre-processing methods are able to successfully approximate the FiND world. Therefore, we utilize an in-processing method that evaluates a model's performance for increasing strengths of a fairness constraint. We observe that for FiND world data as well as for pre-processed data fairness is satisfied without sacrificing predictive performance, confirming that the pre-processing effectively eliminated bias. 
    Concurrently, we show how all fairness metrics that are subject to the trade-off between competing fairness notions are simultaneously satisfied when using pre-processing methods that approximate the FiND world.   
\end{itemize}
Lastly, we validate our findings from the simulated setup on a real-world dataset, using data from the Home Mortgage Disclosure Act (HMDA) dataset.\footnote{Consumer Financial Protection Bureau, \textit{FFIEC HMDA Platform}, \url{https://ffiec.cfpb.gov/}} Our results indicate that the causal pre-processing methods are equally capable of successfully aligning fairness and performance on real datasets while satisfying multiple fairness metrics. We therefore present actionable methods for how both the fairness-accuracy trade-off and the trade-off between various fairness metrics can be overcome in practice.

\subsection{Related Work}
\label{sec:related_work}
While the widespread assumption that there is a trade-off between fairness and accuracy has already been explored from various angles~\cite{plecko_trade_2024, zliobaite_2015, hardt_eo_2016, corbett_cost_2017, cooper_emergent_2021}, a growing number of fairML approaches are challenging this trade-off~\cite{sharma_data_2020, cruz_unpro_2024, rodolfa_empirical_2021, maity_2021, berk_convex_2017}.
In particular, recent works suggest bias as a possible cause of this apparent conflict~\cite{blum_recovering_2020, sharma_how_2024, menon_cost_2018, dutta_is_2020, favier_how_2023, goethals_beyond_2024, fish_confidence-based_2016}. For instance, enforcing fairness constraints has been shown to enhance accuracy when evaluated on unbiased data~\cite{blum_recovering_2020, sharma_how_2024}. Besides, \citet{menon_cost_2018} frame the trade-off in terms of the dependence between the PA and the target label, showing that one can only achieve maximum accuracy and fairness simultaneously in the case of perfect independence. 
As we determine in the causal framework in Section~\ref{sec:group_fairness}, the assumption of (conditional) independence between PA and outcome 
aligns with the assumptions of the FiND world. 

Closely related to our approach is the work by \citet{wick_unlocking_2019} who tackle the fairness-accuracy trade-off from a non-causal viewpoint. They conclude that the trade-off itself is a false notion if accuracy is measured on test data that is equally biased as the training data.
Therefore, one must assume that the corresponding accuracy measurements are also biased and it is crucial to
evaluate accuracy on the unbiased -- fair -- labels. By simulating fair labels on the test set, they show that fairness and accuracy are positively related. We build on these findings by integrating the generation of fair labels into the causal FiND world framework. This allows us to provide a theoretical explanation for the trade-off from a causal perspective and, additionally, to offer a practical solution for overcoming the trade-off by pre-processing that approximates the FiND world. Therefore, we are able to address the trade-off in both simulated and real-world settings.

Regarding the impossibility theorem, \citet{bell_possibility_2023} examine the two exceptions of the impossibility theorem where fairness metrics can be satisfied simultaneously: the case of equal group base rates and the case of perfect prediction. They conceive ``fairness regions'' in which they quantify the possible options of aligning the fairness metrics subject to the trade-off when one of these special cases is approximately satisfied. We also investigate the special case of equal group base rates and derive that it is, by design, fulfilled in the FiND world. 

\section{Theoretical Framework} \label{sec:theoretical_frame}

\subsection{Conception of FiND World} \label{sec:FiND_world}
Fairness in ML has often been approached by employing quantitative group fairness metrics. However, the use of such ``classical'' metrics has been criticized, as their blind application -- without considering the underlying structure of the data or potential biases -- can be blatantly unfair~\cite{dwork_fairness_2011, corbett-davies_measure_2023, mitchell2021algorithmic}. \citet{bothmann_what_2024} therefore derive a philosophically grounded definition of what a normatively fair treatment is and support this by taking into account causal considerations. 

In fact, even from a philosophical perspective, there appears to be no single answer to ``what fairness is''. Instead, they show that the conception of fairness always depends on multilayered normative evaluations that depend on the context or task at hand: Building on Aristotle, a treatment is considered fair if equals are treated equally and unequals unequally. They define the concept of ``task-specific equality'' and thereby require normative stipulations to define when individuals are considered to be equal in a certain task.
When considering causal relationships in the real world, a PA can have a causal effect on the target.  This may be due to a variety of reasons, including biases introduced during the data collection process or historical biases, stemming from historic discrimination against the protected group. 
They argue that taking such causal relationships into account in a prediction process thus can represent a normatively unfair treatment. The reasoning behind this is that societal norms as well as several legal requirements demand not to differentiate between individuals based on
their PAs. 
This implies that individuals are to be considered equal if they only differ in their association with a protected group. In order for a treatment to be normatively fair, one must therefore move the decision-making process to a world without causal effects from the PA on the target.

\subsubsection{The FiND World}
\label{sec:FiND_world_example}
In the ``fictitious, normatively desired'' (FiND) world introduced by \citet{bothmann_what_2024}, the PAs have no causal effect on the target, neither directly nor indirectly.\footnote{They also provide an adjusted version for the case that certain path-specific effects are normatively deemed admissible, c.f.\ their Def. 3.7. For the ease of presentation, we work with the stricter definition of no causal effects in the remainder -- while extensions are straightforward and only depend on the availability of suitable pre-processing methods.} They distinguish this counterfactual FiND world from the real world, in which direct and indirect causal effects from the PAs on the target may exist.

\begin{figure}[h]
    \centering
    \scalebox{0.7}{

\pgfdeclarelayer{background}
\pgfdeclarelayer{foreground}
\pgfsetlayers{background,main,foreground}

\begin{tikzpicture}

\tikzset{fit margins/.style={/tikz/afit/.cd,#1,
    /tikz/.cd,
    inner xsep=\pgfkeysvalueof{/tikz/afit/left}+\pgfkeysvalueof{/tikz/afit/right},
    inner ysep=\pgfkeysvalueof{/tikz/afit/top}+\pgfkeysvalueof{/tikz/afit/bottom},
    xshift=-\pgfkeysvalueof{/tikz/afit/left}+\pgfkeysvalueof{/tikz/afit/right},
    yshift=-\pgfkeysvalueof{/tikz/afit/bottom}+\pgfkeysvalueof{/tikz/afit/top}},
    afit/.cd,left/.initial=2pt,right/.initial=2pt,bottom/.initial=2pt,top/.initial=2pt}

\tikzstyle{surround} = [thick,draw=black,rounded corners=1mm]
\tikzstyle{scalarnode} = [circle, draw, fill=white!11,  
    text width=1.2em, text badly centered, inner sep=2.5pt]
\tikzstyle{scalarnodenoline} = [  fill=white!11, 
    text width=1.2em, text badly centered, inner sep=2.5pt]
\tikzstyle{arrowline} = [draw,color=black, -latex]
\tikzstyle{dashedarrowcurve} = [draw,color=black, dashed, out=100,in=250, -latex]
\tikzstyle{dashedarrowline} = [draw,color=black, dashed,  -latex]

    
    \node [scalarnode, fill=black!20] at (0,0) (A)   {$A$};
    \node [scalarnode, fill=black!20, below of=A, yshift=-7.5mm, xshift=-30mm] (X_D)  {$\!X_D$};
    \node [scalarnode, fill=black!20, below of=A, yshift=-7.5mm] (X_A)  {$\!X_A$};
    \node [scalarnode, fill=black!20, below of=A, yshift=-7.5mm, xshift=30mm] (Y)  {$Y$};
    \node [scalarnode, fill=black!20] at ($(X_D)!0.5!(Y) + (0,-12mm)$) (X_C)  {$X_C$};


    \path [arrowline] (A) to (X_D);
    \path [arrowline] (A) to (X_A);
    \path [arrowline] (A) to (Y);
    \path [arrowline] (X_C) to (X_D);
    \path [arrowline] (X_C) to (X_A);
    \path [arrowline] (X_C) to (Y);
    \path [arrowline] (X_D.south) to  [bend right=80] (Y.south);
    \path [arrowline] (X_A) to (Y);

    \node[surround, fit margins={
    left=16.25mm,right=16.25mm,
    bottom=18.75mm,top=1.25mm},  fit=(A)] (real_world)  {};
    \node[below, yshift=-2.5mm, font=\Large] at (real_world.south) {\emph{(a)} Factual real world \label{cap:real_world}};


    \node [scalarnode, fill=black!20] at (12,0) (A^F)   {$A$};
    \node [scalarnode, below of=A^F, yshift=-7.5mm, xshift=-30mm] (X_D^F)  {$\!X_D^*$};
    \node [scalarnode, below of=A^F, yshift=-7.5mm] (X_A^F)  {$\!X_A^*$};
    \node [scalarnode, below of=A^F, yshift=-7.5mm, xshift=30mm] (Y^F)  {$Y^*$};
    \node [scalarnode, fill=black!20] at ($(X_D^F)!0.5!(Y^F) + (0,-12mm)$) (X_C^F)  {$X_C$};

    \path [arrowline] (X_C^F) to (X_D^F);
    \path [arrowline] (X_C^F) to (X_A^F);
    \path [arrowline] (X_C^F) to (Y^F);
    \path [arrowline] (X_D^F.south) to  [bend right=80] (Y^F.south);
    \path [arrowline] (X_A^F) to (Y^F);

    \node[surround, dashed, fit margins={
    left=16.25mm,right=16.25mm,
    bottom=18.75mm,top=1.25mm},  fit=(A^F)] (FiND_world)  {};
    \node[below, yshift=-2.5mm, font=\Large] at (FiND_world.south) {\emph{(b)} Counterfactual FiND world \label{cap:FiND_world}};


    \node[single arrow, draw=black, fill=white, 
      minimum width = 40pt, single arrow head extend=8pt,
      minimum height=35mm] at ($(real_world.east)!0.5!(FiND_world.west)$) (trans) {\Large Transformation};
    \node[below of=trans] (methods) { e.g.\ FairAdapt~\cite{plecko_fair_2020}, Warping~\cite{bothmann_what_2024}};
    
\end{tikzpicture}
}
    \caption{Assumed causal DAGs of the \emph{(a)} real world and \emph{(b)} FiND world for the credit application example introduced in Section~\ref{sec:FiND_world_example}, with shaded nodes being observed, i.e., accessible for model training.  In the FiND world, the PA $A$ has no children, its descendants according to the causal graph from the real world lack a causal bias from $A$. To distinguish these counterfactual features from their real equivalents, we denote them with an asterisk. For further details on their transformation, we refer to Section~\ref{sec:approximation}.} \label{fig:DAG}
\end{figure}
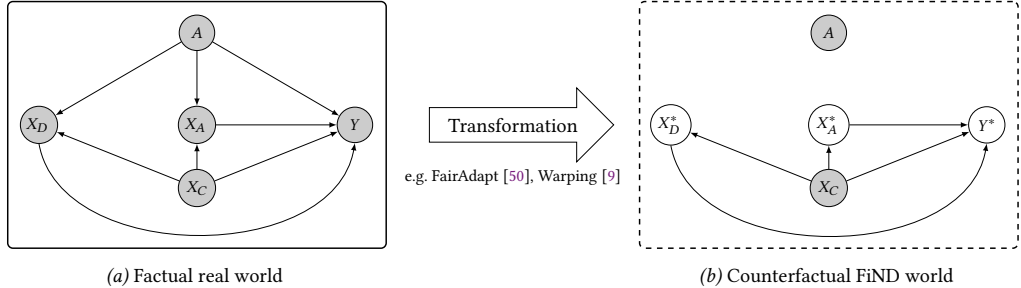

 Assuming acyclic causal relations and causal sufficiency, the model of the FiND world and real world can be represented using a Bayesian network where nodes represent random variables. A directed edge denotes a causal effect, while a directed path with more than two variables implies an indirect one. Throughout the paper we will use the example of a credit application with binary PA $A$, 
 target $Y$ (risk of credit non-repayment), features $X_A$ (credit amount), $X_D$ (debt) and a confounder $X_C$ (age). The corresponding directed acyclic graphs (DAGs) for the real and FiND world are illustrated in \autoref{fig:DAG}. Any PA is assumed to be a root node and is shared between both worlds, i.e., remains unchanged. By contrast, the missing effect of PAs on features yields a counterfactual FiND representation which we denote by $\vX^*$ to distinguish them from $\vX$ that are observed in the real world.

For any $a, a'$ of the set of possible group memberships $\mathcal{A}$, the setting of the FiND world ensures that there is no difference in distributions of outcomes $Y^*$ regarding the protected group $a$ and unprotected group $a'$. Further, individuals of $a$ and $a'$ receive equal outcome $Y^*$ if they are equal with respect to their counterfactual features $\vX^*$ and factual feature(s) $X_C$. In this manner, the FiND world represents a conceptual framework that ensures fairness both on the group level and on the individual level (further explanation follows in Section~\ref{sec:relation_tradeoffs}).

\subsubsection{FiND World vs. ``We're All Equal'' Worldview}
\label{sec:WAE_fairness}
While differing in their philosophical derivation of a fair world, the “We’re All Equal” (WAE) worldview by \citet{friedler_impossibility_2016, friedler_impossibility_2021} shares theoretical implications with the FiND World, but from a non-causal approach. The WAE worldview reflects the assumption that in a construct space -- an idealized, unobservable space holding unbiased information of individuals -- all groups are essentially the same. However, biases arise when mapping from the construct space to the observed space, resulting in discriminatory representations that get transported into the decision space. In order to be fair, the decision-making process must therefore be shifted to the construct space. 
Therefore, WAE demands the target $Y$ to be independent of the PA $A$: 
 \begin{equation}
    Y \perp\!\!\!\perp A.
    \label{eq:independence}
\end{equation}
The FiND world can be transferred to the WAE worldview to the extent that if there exists no causal effect (neither direct nor indirect) from $A$ on $Y$, $Y$ is also statistically independent of $A$. Therefore, (\ref{eq:independence}) is fulfilled in the FiND world. 

\subsubsection{FiND World vs. Counterfactual Fairness}
\label{sec:counterfactual_fairness}
Counterfactual fairness~\cite{kusner_counterfactual_2018} is a causal fairness notion that compares factual and counterfactual predictions. 
In the counterfactual world, an individual with a PA $A = a$ belongs counterfactually to the group $A = a'$. 
A predictor $\hat{Y}$, defined with latent exogenous variables $\vU$ and observable features $\vX$, is considered counterfactually fair iff, for all $\vx \in \mathbb{R}^p$ and $a, a' \in \mathcal{A}$: 
\begin{equation}
P(\hat{Y}_{A \leftarrow a}(\vU) = y \mid \vX = \vx, A = a) = P(\hat{Y}_{A \leftarrow a'}(\vU) = y \mid \vX = \vx, A = a).
\label{eq:counterfactual}
\end{equation}
This ensures that the predicted outcome $\hat{Y}$ remains unchanged when the individual's PA $A$ is counterfactually altered. This condition is fulfilled in the FiND world.\footnote{See~\cite{bothmann_what_2024}, Section 3.4, for a more detailed comparison of the two concepts.} 
In addition, counterfactual fairness does not necessarily fulfill fairness on the group level~\cite{silva_counterfactual_2024}, which is a key difference to the FiND world (see Section~\ref{sec:group_fairness}).

\subsubsection{FiND World vs. Individual Fairness} 
Individual fairness, as proposed by \citet{dwork_fairness_2011}, postulates that similar individuals should be treated similarly. This is formalized using task-specific similarity metrics that ensure that the outcomes for two individuals of different protected groups are similar if their features are similar. However, they acknowledge the difficulty of specifying these metrics and propose to base their choice on the context of the task and feature space. 
The FiND world ties in with this by normatively redefining the``task-specific equality'' on the basis of social and legal requirements. This involves a transformation procedure from the observed real world  -- where dissimilarities based on PAs may exist but are deemed unjust -- to a FiND world -- where these dissimilarities are removed. This means individuals are treated equally if they are equal in the FiND world (and if the predictor is individually well-calibrated, see \citet{bothmann_what_2024}, Def. 3.8) and thereby, individual level fairness is fulfilled. Note that this is generally different from the approach of \citet{dwork_fairness_2011}, where similarity is evaluated in the real world.

\subsection{Relation to Fairness Trade-offs}
\label{sec:relation_tradeoffs}
We can derive relations between the FiND world and the resolution of several existing fairness trade-offs.  A key factor responsible for this lies in the fact that the FiND world also inherently fulfills several group fairness notions. 

\subsubsection{Relation to Group Fairness}
\label{sec:group_fairness}
The assumption of $Y \perp\!\!\!\perp A$ in the FiND world has important implications for the fulfillment of several group metrics, as also highlighted by \citet{loftus_2018}. They make a similar connection between group fairness and counterfactual fairness for cases where there are no causal paths between $Y$ and $A$. This exactly is characterized in the FiND world. 
\paragraph{Demographic Parity}
Perhaps the most common group metric is \emph{demographic parity},
defined as follows:
\begin{equation}
    P(\hat{Y} = y \mid A = a) = P(\hat{Y} = y \mid A = a'),
    \label{eq:dempar}
\end{equation}
for all $y\in\{0,1\}$, $a, a' \in \mathcal{A}$. An alternative way of framing this condition is by the assumption of \emph{Independence}~\cite{barocas-hardt-narayanan}, 
\begin{equation}
   \hat{Y} \perp\!\!\!\perp A. 
   \label{eq:dempar_ind}
\end{equation}
Since $A$ is assumed to be a root node and no back door paths can exist, any statistical dependence between $A$ and $\hat{Y}$ is either due to a causal effect from $A$ on $Y$ or introduced by a common child $X$.
From $A \perp\!\!\!\perp \vX^*,Y^*$ in the FiND world immediately follows that a predictor $\hat{Y}^*$
trained on its representation, i.e., 
$\hat{Y}^* = \hat{f}(\vX^*)$ satisfies demographic parity.

\paragraph{Equalized Odds}
Similar connections apply to \emph{equalized odds}, which requires that both groups have equal error rates. Therefore, it includes both the conditions of \emph{false positive error rate balance} and \emph{false negative error rate balance}.
Formally, equalized odds is defined as:
\begin{equation}
    P(\hat{Y} = y \mid A = a, Y = y) = P(\hat{Y} = y \mid A = a', Y = y),
    \label{eq:eqodds}
\end{equation}
for all $y\in\{0,1\}$, $a, a' \in \mathcal{A}$. 
This condition ensures that the predictor $\hat{Y}$ is conditionally independent of $A$ given the true outcome $Y$, which also falls under the term \emph{Separation}~\cite{barocas-hardt-narayanan} and can be written as:
\begin{equation}
    \hat{Y} \perp\!\!\!\perp A \mid Y.
    \label{eq:eqodds_ind}
\end{equation}
The independence of the target $Y^*$ (\ref{eq:independence}) and the predictor $\hat{Y}^*$ (\ref{eq:dempar_ind}) w.r.t. the PA $A$ in the FiND world directly imply that the predictor satisfies (\ref{eq:eqodds_ind}) by the compositional and weak union axioms of the compositional graphoid induced by the causal DAG~\cite{Lauritzen_composition_2018}. 

\paragraph{Predictive Parity}
In the same way, relations can be drawn to \emph{Predictive Parity}~\footnote{For hard-label predictors, this condition is referred to as \emph{predictive parity}, as it ensures equal positive predictive values (PPV) for $a$ and $a'$. For predicted scores $S$, this condition is commonly termed \emph{calibration}, demanding that for each score $s$ the probability of belonging to the positive class is equal for both $a$ and $a'$}), which ensures that for a predicted outcome $\hat{Y}$ individuals in both protected and unprotected group have equal probability to truly belong to class $Y$. Formally, this requires:
\[
P(Y = y \mid \hat{Y} = y, A = a) = P(Y = y \mid \hat{Y} = y, A = a'),
\]
for all $y\in\{0,1\}$, $a, a' \in \mathcal{A}$.
This condition is also known as \emph{Sufficiency}~\cite{barocas-hardt-narayanan} and implies that the true outcome $Y$ is independent of $A$ when conditioned on the predicted outcome $\hat{Y}$, expressed as:
\begin{equation}
   Y \perp\!\!\!\perp A \mid \hat{Y}.
   \label{eq:calibration}
\end{equation}
Following the same reasoning as for equalized odds, (\ref{eq:calibration}) is a direct consequence of $Y^* \perp\!\!\!\perp A$, leading to predictive parity being inherently satisfied in the FiND world.

\subsubsection{Resolving the Trade-off Between Group Fairness Metrics} 
\label{eq:trade_off_group_fairneess}
Parts of the fairML literature surround the problem that several group fairness metrics are mathematically incompatible with each other. This is known as the \emph{impossibility theorem}~\cite{kleinberg_inherent_2017, chouldechova_fair_2017}, which states that false positive rate balance, false negative rate balance, and predictive parity cannot be satisfied simultaneously
unless for two special cases: one must either have equal group base rates (also falls under the term \emph{prevalence}) for the protected and unprotected group or perfect prediction.

However, as we have just derived, these metrics are all inherently fulfilled in the FiND word.  In fact, the FiND world represents one of the special cases of the impossibility theorem, namely equal base rates among groups. This is due to the independence assumption of $Y \perp\!\!\!\perp A$ of the FiND world, which implies that individuals from protected and unprotected groups have the same probability of belonging to the positive class:
\begin{equation}
    P(Y=1|A=a)=P(Y=1|A=a')=P(Y=1).  
\end{equation}
 This assumption can be normatively justified if we presume that different group base rates in the real world are entirely attributable to historical discrimination and biases stemming from the data collection process. 
Therefore, the FiND world is able to overcome the trade-off between competing fairness metrics naturally by design.  

\subsubsection{Resolving the Conflict Between Group Fairness and Individual Fairness} 
Additionally, as the FiND world approaches fairness on the individual level, it also overcomes a conflict between group fairness and individual fairness notions. A similar observation is made by \citet{binns_apparent_2019}, who argue that the conflict between individual and group fairness depends on the worldview and the underlying normative principles a decision-maker considers. By, e.g., adopting a WAE worldview, both individual and group approaches are driven by the same moral assumptions that groups should be treated equally based on their PAs. The same considerations apply to the FiND world, as observed group differences are assumed to be due to biases in the real world and should therefore not be taken into account.  

Coming back to the financial lending example, the existence of discrimination based on race and gender has repeatedly been observed in various US mortgage markets \cite{alg_lee_2021}. From a policy perspective, to base the approval of a loan on such historical disparities would constitute an unjust practice and is legally restricted under, e.g., the Fair Housing Act and the Home Mortgage Disclosure Act (HMDA). This raises the need for decision-makers (in this scenario, banks) to proactively implement \emph{substantive equality} as described by \citet{wachter_bias_2021}. They argue that in order to achieve fair treatment, one has to account
for historical inequalities by actively re-leveling discriminated individuals via \emph{bias-transforming} actions (which can also be understood as some type of \emph{affirmative action}). This can be connected to the transformation from the real world to FiND world. 

\subsubsection{Resolving the Fairness-Accuracy Trade-off}
As we move the prediction process into the FiND world, we are also able to overcome the fairness-accuracy trade-off. Since all the investigated fairness notions are inherently embedded within the FiND world, enforcing a fairness metric no longer leads to a decrease in predictive performance; instead, fairness and predictive performance become aligned. Furthermore, this alignment enables us to reframe the quest for fair models: We can focus on high predictive performance using the data representation of the FiND world for which fairness naturally holds. 

The FiND world thus resolves both fairness trade-offs by providing a unified causal framework. Moreover, our findings have significant practical implications: if we can approximate the FiND world through suitable pre-processing methods, it becomes possible to overcome these trade-offs in real-world applications. Successfully doing so would eliminate the need to explicitly enforce specific fairness metrics, allowing high predictive performance to naturally serve as the primary notion of fairness.

\section{Approximating the FiND World} 

\subsection{Pre-processing} \label{sec:transformation}
Since we do not have access to data from the FiND world in practice, we need to project the real-world data into the FiND world as a preliminary step. This can be understood as some type of \emph{fair representation learning} \cite{zemel_learn_2013}. Various methods have been proposed to learn fair representations of the data (e.g., \cite{zemel_learn_2013, calmon_optimized_2017, ruoss_learn_2020, lahoti_ifair_2019, xu_fair_2024}). However, most of them present non-causal techniques and do not apply to the specific assumptions of the FiND world. 
Against this background, we consider two causal pre-processing techniques to approximate the FiND world. 

\paragraph{Fairadapt}
With fairadapt,  ~\citet{plecko_fairadapt_2021-1} present a pre-processing method using a causal approach. Their method is based on quantile preservation and uses quantile regression forests \citep{meinshausen_quantile_nodate}. They aim to approximate a counterfactual world by constructing ``fair twins''. Their method constructs these fair twins by transforming the observational distribution for each individual to a fair-projection distribution. In doing so, they demand all individuals to have the same PA after the fair twin projection, thereby setting the protected group to its baseline value.  
Note that this differs slightly from the warping method (see below), which does not intervene on the PA but only on its descendants. 
However, this distinction is not a problem here, as the protected attribute from the real world can replace the baseline value in the joint distribution after the projection of the features $\vX$ to meet the definition of the FiND world. 
We use their R package \texttt{fairadapt}\footnote{\url{https://cran.r-project.org/web/packages/fairadapt/index.html}} to adapt the train and the test dataset, and in the following refer to this as the ``adapted world'' data. 
\paragraph{Residual-based Warping}
\citet{bothmann_causal_2024} propose another causal approach by introducing a ``residual-based warping'' method to approximate the FiND world. Their method consists of intervening on the paths from the PA to its descendants by transforming the observations for the protected group to the corresponding quantile of the unprotected group's distribution. They reduce the problem of estimating full distributions to estimating models for the location parameters of their distributions. Therefore, they derive individual probability ranks by using a residual-based approach. The method thereby derives ``rank-preserving interventional distributions'' (RPID), i.e., individuals of the protected group maintain the group-specific pre-intervention ranks for each warped variable. By symmetry, reversing the warping direction from the unprotected to the protected group is also feasible, as it produces the same result of eliminating the effect of the PA. 
We use R functions 
from their GitHub repository\footnote{\url{https://github.com/slds-lmu/paper_2024_rpid}} 
for warping training and test data, which will be referred to as the ``warped world'' data.

\subsection{In-processing}
\label{sec:in-processing}
Additionally,  we use an in-processing method to evaluate both the trade-off between fairness and performance on different datasets and whether the pre-processing is able to approximate the FiND world successfully. In-processing methods generally work by enforcing a desired
fairness constraint during model training, see, e.g.,  \cite{zafar_fairness_2017, zafar_fairness_2019, cotter_opt_2019, cruz_fairgbm_2023, celis_2019, donini_emp_2018}. In particular, we are interested in whether for a model trained on i.i.d data from the real world, ${\Dtrain \coloneq \{ ( \vx_i, y_i ) \}_{i=1}^N}$\,, and evaluated on pre-processed data, the performance improves when enforcing a fairness constraint. 
For this purpose, we consider the following constraint optimization problem, which we implement by adding a fairness regularization term to the empirical risk:

\paragraph{Regularized Empirical Risk}
For individuals $i \in \{1,2, \dots, N\}$ in $\Dtest$, their corresponding targets $y_i \in \{0, 1\}$ and features $\vx_i \in \mathbb{R}^d$, we obtain the regularized empirical risk
\begin{equation}
    R_\text{reg}\left(f, \Dtrain\right) = \sum_{i=1}^{N} L\big(y_i, f(\vx_i)\big) + \lambda \cdot C(\pitrain),
    \label{eq:emp_risk}
\end{equation}
where $L\big(y_i, f(\vx_i)\big)$ denotes the Bernoulli loss with predicted log-odds $f(\vx_i)$, and $\pitrain$ the vector of predicted probabilities for all data samples $\vx_i \in \xtrain=\{\vx_i\}_{i=1}^N$ with
\begin{align}
   \hat{\pi}_i \coloneq
   \sigma\big(f(\vx_i)\big) =
   \frac{1}{1 + e^{-f(\vx_i)}}.\label{eq:predictions}
\end{align}
The regularization term $C(\vpi)$ represents the fairness constraint controlled by the penalty parameter $\lambda$. We refer to $C(\vpi)$ as a regularization term, yet its purpose is not to reduce model complexity or improve generalization, but to 
favor models that lead to fairer predictions.

\paragraph{Fairness Constraint $C$}
We penalize the difference in the average predicted probabilities between the protected group memberships $a$ and $a'$ with
\begin{equation}
    C(\vpi) \coloneq \left| \frac{1}{N_{a}} \sum_{i:\,A_i = a} \hat{\pi}_i  - \frac{1}{N_{a'}} \sum_{j:\,A_j = a'} \hat{\pi}_j\right|,
    \label{eq:fairness_constraint}
\end{equation}
where $N_a$ and $N_{a'}$ denote the total number of individuals per group. This can be seen as a form of demographic parity constraint~\cite{agarwal_reductions_2018, konstantinov_fairness_2021} that is a necessary condition of the FiND world as we proved in Section~\ref{sec:group_fairness}.

\paragraph{Gradient Boosted Trees}
To learn our model, we use gradient boosted trees in the R implementation of \texttt{xgboost} \citep{chen_xgboost_2014}. We split our data randomly into 80\% training data $\Dtrain$ and 20\% test data $\Dtest \coloneq \left\{ (\vx_i, y_i) \right\}_{i = N+1}^M$
and determine optimal hyperparameters $d$ (depth of the trees) and $\eta$ (learning rate) via random search with 3-fold cross-validation on $\Dtrain$. 

\paragraph{Fairness-Accuracy Trade-off Evaluation}
To assess whether the pre-processing is able to resolve the fairness-accuracy trade-off, we investigate whether the performance of a model evaluated on pre-processed data improves as the fairness constraint $C$ increases.
As a preliminary step, we find the smallest penalty parameter $\lambda^*$ for which ${C\big(f(x)\big) < \epsilon}$ is fulfilled
via grid search 
(trained on $\Dtrain$ with the above optimal hyperparameters and evaluated on $\Dtest$)
.
Since $\epsilon = 0$ is unfeasible for probabilistic classifiers (for non-trivial cases such as $\hat{\pi}= 1.0$ or $\hat{\pi}= 0.0$), we set $\epsilon = 0.01$, thereby allowing fairness differences of up to $1\%$

\begin{algorithm}[ht]
\caption{Fairness-Accuracy Trade-off Evaluation}
\label{alg:fairness-eval}

\begin{algorithmic}
\Require Training dataset $\Dtrain$, test dataset $\Dtest$, learning rate $\eta$, tree depth $d$, optimal $\lambda^*$, interpolation steps $S$
\Ensure AUC values $\mathbf{p}$, empirical group disparities $\vdelta$

\For{$n \gets 0, 1, \dots, S+1$}
    \State $w \gets \frac{n}{S+1}$
    \State 
    $\displaystyle
    \hat{f} \gets \underset{f \in \mathcal{H}}{\arg\min} \sum_{x_i   \in \Dtrain} L\big(y_i, f(\vx_i)\big) \;+\; w \cdot \lambda^* \cdot C\big(\pitrain\big)
    $
    \Comment Train model $\hat{f}$ with parameters $\eta$, $d$ with $C$ from \eqref{eq:fairness_constraint}
    \For{$\vx_i \in \xtest$}
        \State $\hat{\pi}_i \leftarrow \sigma\big(\hat{f}(x_i)\big)$
        \Comment Predict probabilities for test features $\xtest \coloneq \Dtest\setminus \ytest$
    \EndFor
    \State $\hat{\delta}_n \gets C(\pitest)$
    \Comment Evaluate fairness via group disparity defined in \eqref{eq:fairness_constraint}
    \State $\displaystyle
    p_n \leftarrow \text{auc}\big(\pitest, \ytest\big)
    $
    \Comment Evaluate performance via AUC
\EndFor

\end{algorithmic}
\end{algorithm}

We then apply Algorithm~\ref{alg:fairness-eval} that trains and evaluates models for different penalty parameters $\lambda=w \lambda^*$, where  $w$ ranges from 0.0 to 1.0 in increments of $\frac{1}{S+1}$, i.e., interpolates between the unconstrained model and the one with an $(1-\epsilon)$-fulfillment of the fairness constraint $C$. In our experiments, we set the number of interpolation steps to $S=9$ leading to 11 models in total. To evaluate the relation between fairness and accuracy, we record for each model its fulfillment of fairness by measuring the difference in predicted group probabilities and evaluate performance using the area under the ROC curve (AUC). We choose AUC over accuracy as the performance measure, since it allows for a more nuanced assessment when classes are imbalanced.

\section{Resolving the Trade-offs via Pre-Processing}

\subsection{Testing Approximation to the FiND World}
\label{sec:approximation}

The approximation of the FiND world by the two pre-processing methods can be evaluated using Algorithm~\ref{alg:fairness-eval}. This is based on the following rationale: We have already derived in Section~\ref{sec:group_fairness} that in the FiND world demographic parity (DP) holds. Thus, enforcing the fairness constraint  $C$ during model training using real world training data $\Dtrain$ should increase performance on FiND world test data $\Dtest^*$ , as the FiND world accurately reflects the enforced fairness. If for pre-processed test data we observe this same relationship -- namely the same increase in performance -- we can conclude that the pre-processed data was able to learn the representation of the FiND world. 
In the following, we investigate whether the fairadapt and warping pre-processing methods are able to approximate the FiND world and can therefore overcome both the fairness-accuracy trade-off and the trade-off between several fairness metrics. To further validate the successful approximation of the FiND world by the pre-processing methods, we additionally check whether the differences in distributions of transformed variables between groups were resolved (see \ref{app:approx_sim} and \ref{app:approx_HMDA}).

\paragraph{Simulation Setup}
We return to the credit application example introduced in Section~\ref{sec:FiND_world_example}. According to the corresponding DAG depicted in \autoref{fig:DAG}, we generate synthetic data for both worlds.
We consider the PA to be binary and equally distributed, i.e.,\ $\pi_a=50\%$ (this can, e.g., mimic gender). The simulated real world model contains direct causal effects of the PA $A$ on the features $\vX$ and target $Y$, while the FiND world model has no causal path from $A$ to neither $\vX$ nor $Y$. For more details on the setup, see Appendix \ref{app_sim_setup}. In total, we conduct 25 simulation runs that each produces a dataset for the real and FiND world.  
We then apply the fairadapt and the warping pre-processing to all simulated real world datasets. 

\subsection{Resolving the Trade-off Between Fairness and Performance}
\label{sec:sim_fairness_performance}

\begin{figure}[h]
    \centering
    \begin{subfigure}[b]{0.23\textwidth}
        \includegraphics[width=\linewidth]{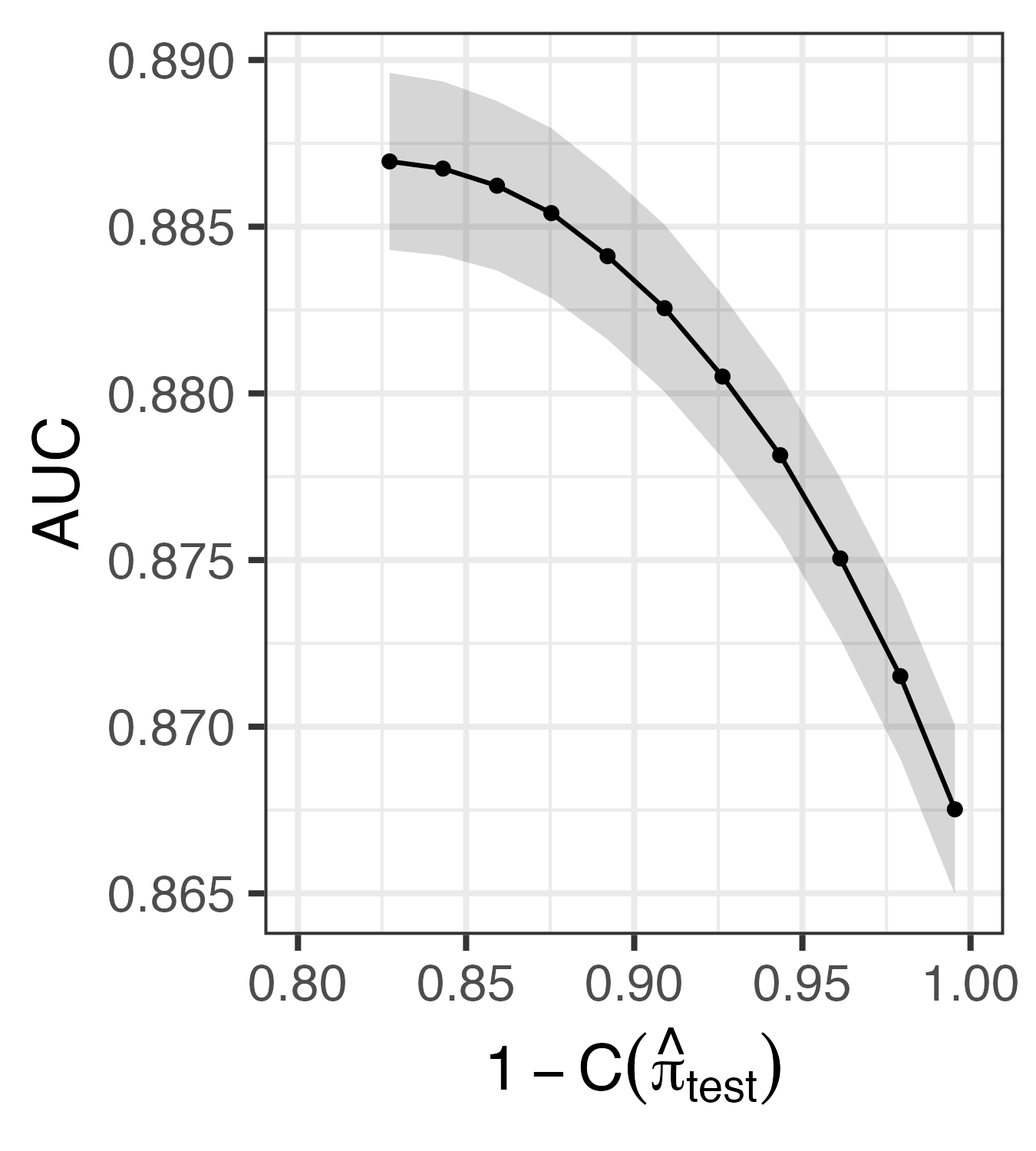}
        \caption{Real world $\Dtest$}
        \label{fig:real_tradeoff}
    \end{subfigure}
    \hfill
    \begin{subfigure}[b]{0.23\textwidth}
        \includegraphics[width=\linewidth]{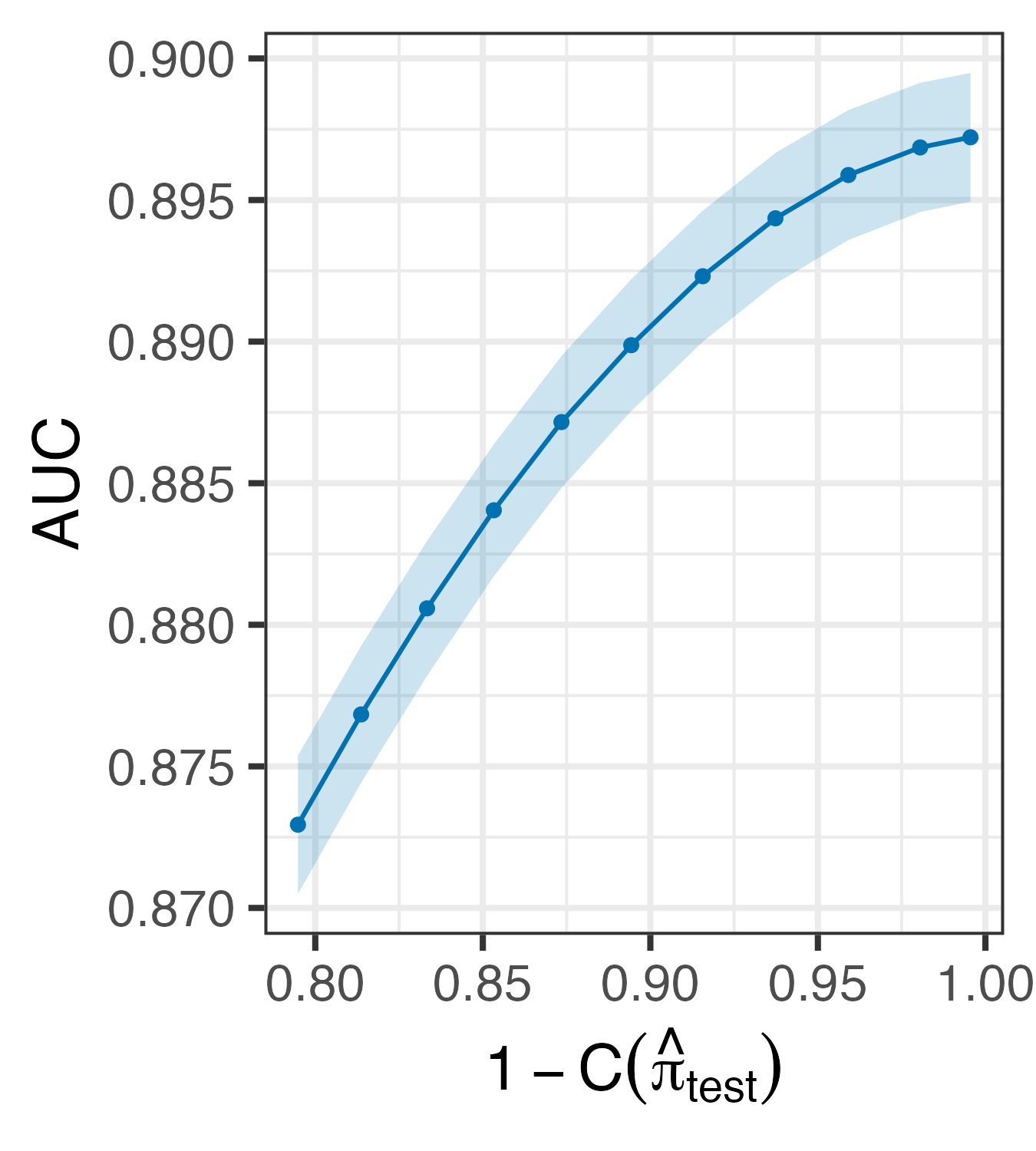}
        \caption{FiND world $\Dtest^*$}
        \label{fig:find_tradeoff}
    \end{subfigure}
    \hfill
    \begin{subfigure}[b]{0.23\textwidth}
        \includegraphics[width=\linewidth]{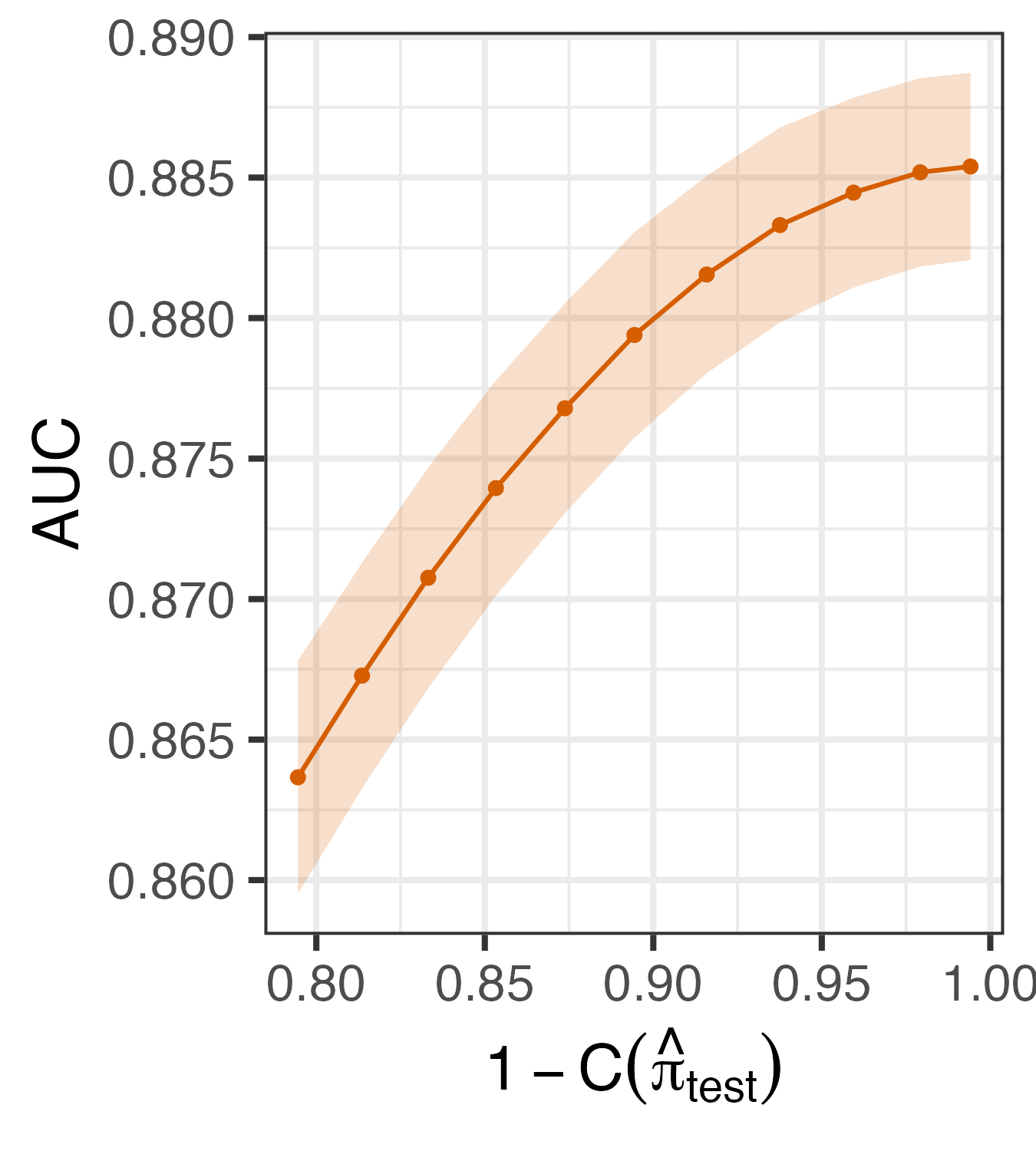}
        \caption{Adapted world $\Dtest^\text{a}$}
        \label{fig:adapt_tradeoff}
    \end{subfigure}
    \hfill
    \begin{subfigure}[b]{0.23\textwidth}
        \includegraphics[width=\linewidth]{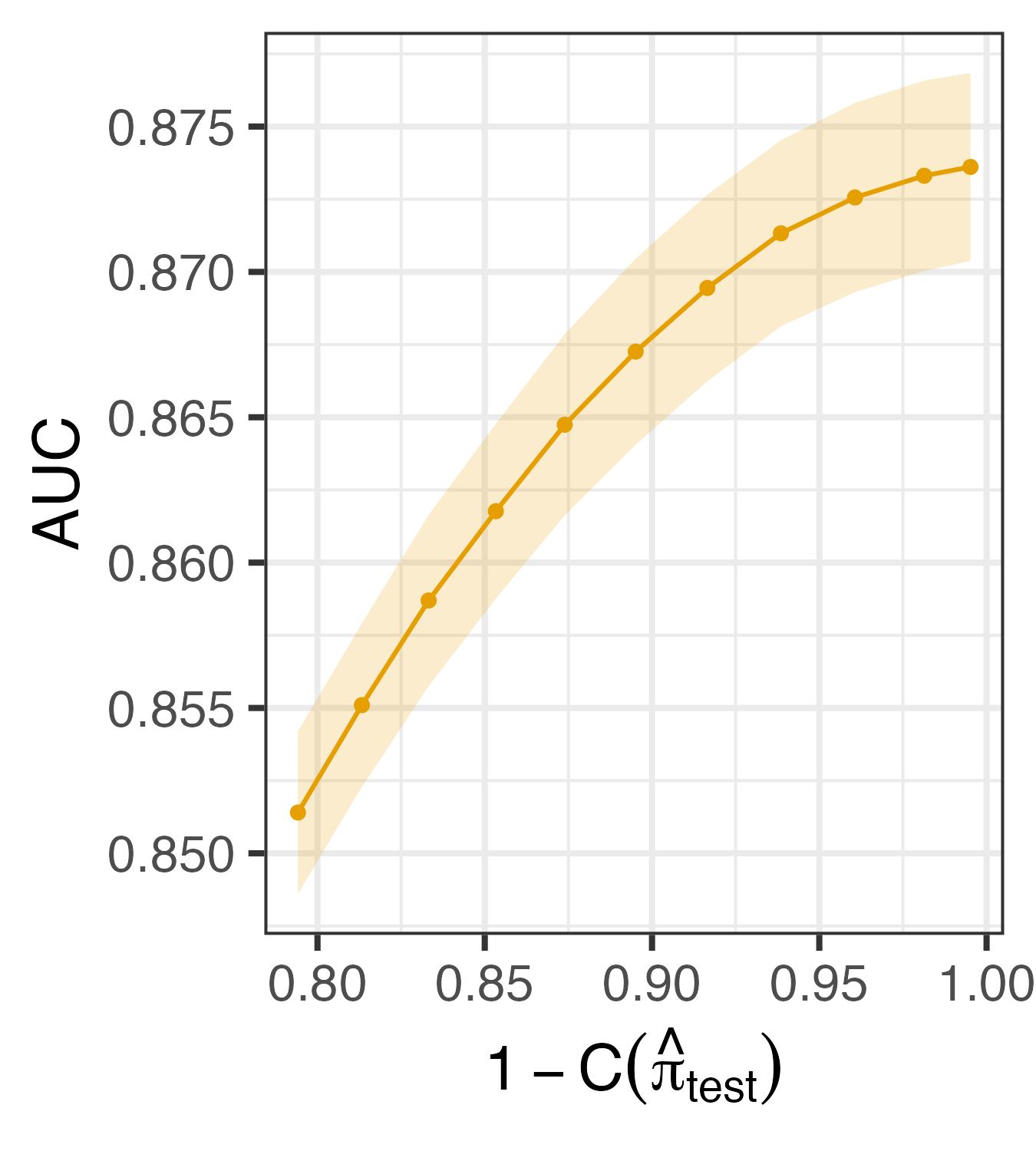}
        \caption{Warped world $\Dtest^\text{w}$}
        \label{fig:warp_tradeoff}
    \end{subfigure}
    \caption{Fairness-performance curves of predictors trained on real-world data $\Dtrain$ for the simulation study in Section~\ref{sec:sim_fairness_performance}, with 95\% confidence intervals. \emph{(a)} Real world test data  does not approximate \emph{(b)} the FiND world due to inherent bias. By contrast, \emph{(c)} the adapted world and \emph{(d)} warped world data do approximate the FiND world, as increasing fairness leads to increased AUC (same monotonic relation as for \emph{(b}) FiND world data .}
    \label{fig:approx_find_sim}
\end{figure}

\autoref{fig:approx_find_sim} presents the fairness-accuracy trade-offs in the different worlds. Using Algorithm~\ref{alg:fairness-eval}, we train models for increasing weights of fairness on real world $\Dtrain$, but evaluate them regarding performance (AUC) and fairness on real world, FiND world, adapted world, and warped world test data, respectively. We use $1 - C(\pitest)$ as a fairness measure for better 
interpretability (high value is better). We additionally record the $95\%$~confidence intervals of the AUC.

 For models trained and tested on real world data, we observe the typical trade-off between fairness and performance (\autoref{fig:real_tradeoff}). 
 In \autoref{fig:find_tradeoff}, we evaluate the same set of classifiers trained on real world data, but this time measure performance on FiND world $\Dtest^*$. We see the exact opposite pattern: Classifiers with 
 low fairness violation $C(\pitest)$ are also more accurate. This also aligns with our theory: When evaluating models that enforce fair predictions (in terms of DP) on data that corresponds to the FiND world where fairness is naturally fulfilled, a model that is ``fairer'' will also be more accurate.  

 We observe the same relationship between fairness and performance also for adapted world  (\autoref{fig:adapt_tradeoff}) and warped world (\autoref{fig:warp_tradeoff}) test data. On both pre-processed test sets, the performance of models increases with lower fairness violations. 
 We conclude that the pre-processed data are equally capable of reflecting a world that incorporates fairness in the sense that there are no differences between groups in their outcomes and thus, $Y \perp\!\!\!\perp A$. Consequently, there can be no direct or indirect causal effects from PA $A$ on $Y$. 
 Therefore, we conclude that both pre-processing methods are able to overcome the trade-off between fairness and performance in the simulated setting and were successful in approximating the FiND world.

\subsection{Resolving the Trade-off Between Fairness Metrics} \label{sec:fairness_fairness}
The results from Section~\ref{sec:sim_fairness_performance} offer another important insight. When the pre-processing methods are successful in approximating the FiND world, we do not have to rely on using the in-processing method anymore to achieve fairness by enforcing a fairness constraint. Instead, we can directly train models on pre-processed data that reflects the FiND world in order to achieve fair predictions. 

In the following, we train unconstrained models on the adapted world and warped world datasets, evaluating them in terms of the compatibility of various fairness metrics. We record the fulfillment of fairness regarding demographic parity~(DP), false positive rate balance~(FPR), false negative rate balance~(FNR) and predictive parity~(PPV) between the protected $a$ and unprotected group $a'$ (averaged over 25 simulations). Regarding performance, we report the average AUC. To allow for comparative analyses, we perform the same evaluations on the real world and FiND world datasets. The results are listed in \autoref{tab:imp_res}.

\begin{table}[h]
\caption{Comparison of fairness and performance metrics in real, adapted, and warped world for \emph{(a)} the simulation study to evaluate the pre-processing methods (Section~\ref{sec:approximation})  and \emph{(b)} the HMDA experiment (Section~\ref{sec:validation}). For \emph{(a)} the simulation study, we additionally record the standard deviation (sd) over the iterations in the subscript. For confidence intervals, see \autoref{tab:ci_sim} in \ref{app:approx_sim}.}
\small
\setlength{\tabcolsep}{1.5pt} 
\begin{subtable}{0.58\textwidth}
\caption{Simulation Study}
\begin{tabular}{lccccc}
\toprule
\multirow{2}{*}{World} & \multicolumn{4}{c}{Fairness} & \multicolumn{1}{c}{Performance} \\
& $\text{DP}_\text{sd}$ & $\text{FPR}_\text{sd}$ & $\text{FNR}_\text{sd}$ & $\text{PPV}_\text{sd}$ & $\text{AUC}_\text{sd}$ \\
\midrule
Real & 0.825$_{0.020}$ & 0.782$_{0.024}$ & 0.954$_{0.017}$ & 0.986$_{0.014}$ & 0.887$_{0.007}$\\
FiND & 0.987$_{0.011}$ & 0.989$_{0.011}$ & 0.991$_{0.007}$ & 0.988$_{0.012}$ & 0.897$_{0.006}$\\
Adapted & 0.982$_{0.012}$ & 0.972$_{0.016}$ & 0.984$_{0.013}$ & 0.975$_{0.015}$ & 0.886$_{0.008}$\\
Warped & 0.982$_{0.011}$ & 0.964$_{0.015}$ & 0.974$_{0.016}$ & 0.971$_{0.018}$ & 0.893$_{0.013}$\\
\bottomrule
\end{tabular}
\label{tab:imp_res}
\end{subtable}
\hspace{0pt}
\begin{subtable}{0.35\textwidth}
    \caption{HMDA Experiment}
    \begin{tabular}{lcccccc}
    \toprule
    \multirow{2}{*}{World} & \multicolumn{4}{c}{Fairness} & \multicolumn{1}{c}{Performance} \\
    & DP & FPR & FNR & PPV & AUC \\
    \midrule
    Real & 0.809 & 0.840 & 0.823 & 0.891 & 0.716 \\
    FiND & --- & \multicolumn{2}{c}{---\,unkown\,---} & --- & --- \\
    Adapted & 0.989 & 0.991 & 0.989 & 0.990 & 0.712 \\
    Warped & 0.994 & 0.988 & 0.990 & 0.975 & 0.713 \\
    \bottomrule
    \end{tabular}
    \label{tab:imp_res2}
\end{subtable}
\end{table}

In the real world, we observe large fairness differences for DP and FPR up to $22\%$. In the FiND world, we obtain much higher fairness values, the empirical differences are less than $1.3\%$. This matches our theoretical results that in the FiND world the concerned fairness metrics are inherently satisfied. The same pattern can be validated for the adapted and warped world, although we observe slightly higher differences than in the FiND world of up to
 $3.6\%$. However, this is still considerably lower than in the real world. We therefore consider both pre-processed worlds to approximately fulfill all fairness metrics. From this, we conclude that in our simulated setting the fairadapt and warping pre-processing methods are both able to successfully overcome the trade-off between several fairness metrics. We also conclude that models of real, FiND, adapt and warped world are all equally well performing (where FiND, adapted and warped have even slightly better AUCs than the real world model). However, note that the performance of the real world model is not directly interpretable as it still inherits bias. 

\section{Real Data Experiments}
\label{sec:validation}

To further validate our results, we demonstrate that our findings from the simulated setup can be transferred
to real data experiments for which the FiND world model is inaccessible and needs to be approximated. For this purpose, we use the 2022 Home Mortgage Disclosure Act (HMDA) dataset for the state Wisconsin.\footnote{The dataset is accessible at \url{https://ffiec.cfpb.gov/data-browser/data/2022?category=states&items=WI}.} This dataset is similar to our previously considered credit assessment example and contains information on housing loan applications, including details about the applicants and the corresponding loan decisions.

We consider race as the binary PA $A$. We filter the data regarding race by only selecting Black (protected group $a$) and non-Hispanic White borrowers (unprotected group $a'$), as previous mortgage studies have highlighted the most significant disparities in approval rates and treatment between these two groups~\cite{alg_lee_2021}. This results in a total of 83,808 observations. The original dataset encompasses over 100 variables, for simplicity, we focus on the subset of the following six:
We include loan amount ($X_A$), purpose of the loan ($X_P$), and debt ratio ($X_D$) as features as well as age and gender ($\vX_C$) jointly as confounders. The target $Y$ indicates whether an applicant was granted a loan (1) or not (0)\footnote{Note that the target $Y$ in this case indicates the loan approval and does not contain information about whether the individual repaid the loan or not. This varies slightly from the previous credit application example, where $Y$ denoted the risk of credit repayment.}. For more details on the data setup, see Appendix \ref{app_ex_setup}. 
\autoref{fig:DAG_hmda} shows the assumed DAGs.
The final dataset consists of $94.8\%$ White vs. $5.2\%$ Black applicants and has an overall loan approval base rate of $66.8\%$. The group-specific base rate for White applicants is $67.9\%$, and for Black applicants $48.6\%$.

\begin{figure}[h]
    \centering
    \scalebox{0.7}{

\pgfdeclarelayer{background}
\pgfdeclarelayer{foreground}
\pgfsetlayers{background,main,foreground}

\begin{tikzpicture}

\tikzset{fit margins/.style={/tikz/afit/.cd,#1,
    /tikz/.cd,
    inner xsep=\pgfkeysvalueof{/tikz/afit/left}+\pgfkeysvalueof{/tikz/afit/right},
    inner ysep=\pgfkeysvalueof{/tikz/afit/top}+\pgfkeysvalueof{/tikz/afit/bottom},
    xshift=-\pgfkeysvalueof{/tikz/afit/left}+\pgfkeysvalueof{/tikz/afit/right},
    yshift=-\pgfkeysvalueof{/tikz/afit/bottom}+\pgfkeysvalueof{/tikz/afit/top}},
    afit/.cd,left/.initial=2pt,right/.initial=2pt,bottom/.initial=2pt,top/.initial=2pt}

\tikzstyle{surround} = [thick,draw=black,rounded corners=1mm]
\tikzstyle{scalarnode} = [circle, draw, fill=white!11,  
    text width=1.2em, text badly centered, inner sep=2.5pt]
\tikzstyle{scalarnodenoline} = [  fill=white!11, 
    text width=1.2em, text badly centered, inner sep=2.5pt]
\tikzstyle{arrowline} = [draw,color=black, -latex]
\tikzstyle{dashedarrowcurve} = [draw,color=black, dashed, out=100,in=250, -latex]
\tikzstyle{dashedarrowline} = [draw,color=black, dashed,  -latex]

    
    \node [scalarnode, fill=black!20] at (0,0) (X_C)   {$X_C$};
    \node [scalarnode, fill=black!20, below of=X_C, yshift=-10mm, xshift=-20mm] (X_P)  {$\!X_P$};
    \node [scalarnode, fill=black!20, below of=X_C, yshift=-10mm] (X_D)  {$\!X_D$};
    \node [scalarnode, fill=black!20, below of=X_C, yshift=-10mm, xshift=20mm] (X_A)  {$\!X_A$};
    \node [scalarnode, fill=black!20, below of=X_C, yshift=-10mm, xshift=40mm] (Y)  {$Y$};
    \node [scalarnode, fill=black!20, above of=X_A, yshift=10mm] (A)  {$A$};

    \path [arrowline] (A) to (X_P);
    \path [arrowline] (A) to (X_D);
    \path [arrowline] (A) to (X_A);
    \path [arrowline] (A) to (Y);
    \path [arrowline] (X_C) to (X_P);
    \path [arrowline] (X_C) to (X_D);
    \path [arrowline] (X_C) to (X_A);
    \path [arrowline] (X_P.south east) to [bend right=35](Y.south);
    \path [arrowline] (X_D.south east) to  [bend right=25] (Y.south west);
    \path [arrowline] (X_A) to (Y);

    \node[surround, fit margins={
    left=21.25mm,right=11.25mm,
    bottom=17.5mm,top=2.5mm},  fit=(A)] (real_world)  {};
    \node[below, yshift=-2.5mm, font=\Large] at (real_world.south) {\emph{(a)} Observed real world \label{cap:real_world_HMDA}};


    \node [scalarnode, fill=black!20] at (12,0) (X_C^F)   {$X_C$};
    \node [scalarnode, below of=X_C^F, yshift=-10mm, xshift=-20mm] (X_P^F)  {$\!X_P^\circ$};
    \node [scalarnode, below of=X_C^F, yshift=-10mm] (X_D^F)  {$\!X_D^\circ$};
    \node [scalarnode, below of=X_C^F, yshift=-10mm, xshift=20mm] (X_A^F)  {$\!X_A^\circ$};
    \node [scalarnode, below of=X_C^F, yshift=-10mm, xshift=40mm] (Y^F)  {$Y^\circ$};
    \node [scalarnode, fill=black!20, above of=X_A^F, yshift=10mm](A^F)  {$A$};

    \path [arrowline] (X_C^F) to (X_P^F);
    \path [arrowline] (X_C^F) to (X_D^F);
    \path [arrowline] (X_C^F) to (X_A^F);
    \path [arrowline] (X_P^F.south east) to [bend right=35](Y^F.south);
    \path [arrowline] (X_D^F.south east) to  [bend right=25] (Y^F.south west);
    \path [arrowline] (X_A^F) to (Y^F);

    \node[surround, dotted, fit margins={
    left=21.25mm,right=11.25mm,
    bottom=17.5mm,top=2.5mm},  fit=(A^F)] (FiND_world)  {};
    \node[below, yshift=-2.5mm, font=\Large] at (FiND_world.south) {\emph{(b)} Approximated FiND worlds \label{cap:FiND_world:HDMA}};
    \node[above left] at (FiND_world.south east) {$\circ \in \{\text{a},\text{w}\}$};


    \node[single arrow, draw=black, fill=white, minimum width = 6pt,
    single arrow head extend=4pt,
    minimum height=10mm]
    at ($(real_world.east) + (7.5mm, 10mm)$) (real_arrow) {};
    \node[right of=real_arrow, xshift=6.5mm] (real_data) {$\Dtrain, \Dtest$};

    \node[single arrow, rotate=180, draw=black, fill=white, minimum width = 6pt,
    single arrow head extend=4pt,
    minimum height=10mm] at ($(FiND_world.west) + (-7.5mm, 0)$) (adapt_arrow) {};
    \node[left of=adapt_arrow, xshift=-6.5mm] (adapted_data) {$\Dtrain^\text{a}, \Dtest^\text{a}$};

    \node[single arrow, rotate=180, draw=black, fill=white, minimum width = 6pt,
    single arrow head extend=4pt,
    minimum height=10mm] at ($(FiND_world.west) + (-7.5mm, -10mm)$) (warpe_arrow) {};
    \node[left of=warpe_arrow, xshift=-6.5mm] (warped_data) {$\Dtrain^\text{w}, \Dtest^\text{w}$};
    
\end{tikzpicture}
}
    \caption{Assumed causal DAGs for the HMDA dataset of \emph{(a)} the real and \emph{(b)} the approximated FiND world alongside their corresponding datasets. The non-shaded nodes in the approximated worlds are transformed since they are descendants of the PA $A$ in the real world. They are denoted with either \text{a} or \text{w} to indicate the applied transformation, FairAdapt~\cite{plecko_fair_2020} or Warping~\cite{bothmann_what_2024}.}
    \label{fig:DAG_hmda}
\end{figure}
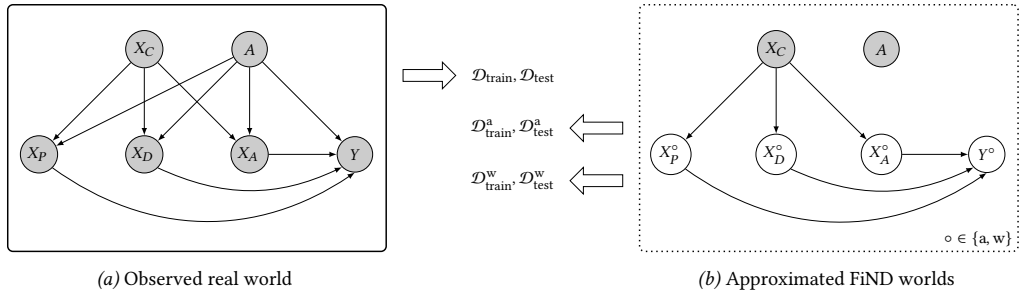

\subsection{Resolving the Trade-off Between Fairness and Performance}
As we do not have access to the FiND world in this real data setting, we rely solely on pre-processing in order to obtain unbiased data that reflects the FiND world. For this purpose, we apply the fairadapt and the warping pre-processing methods on the real HMDA train and test data. To check if the pre-processing successfully eliminated all causal effects from the PA $A$ on $Y$ and is thereby able to approximate the FiND world, we again apply the in-processing Algorithm \autoref{alg:fairness-eval}. The results are displayed in \autoref{fig:comparison_tests}. Additionally, we compare distributions in Appendix \ref{app:approx_HMDA}.

For the real world HMDA data in \autoref{fig:hmda_app_real}, we again observe the common trade-off between fairness and performance, where an increase in fairness from $80\%$ (fairness of the unconstraint model) to $100\%$ is accompanied by a decrease in performance. 
The absolute decrease in performance is rather small in this case, which is due to the highly imbalanced classes of the PA where $a'$ only makes up $5\%$ of the data. 
In \autoref{fig:hmda_app_adapt} and \autoref{fig:hmda_app_warped}, we again see a positive connection between fairness and performance. Models that enforce higher fairness constraints during training on real world train data also achieve higher performance on pre-processed test data. 
 On both pre-processed test sets, we see a range in fairness increase from $85\%$ (unconstraint model) up to $100\%$ which is accompanied by an increase in performance. 
 This increase is again rather small in absolute values but nonetheless indicates that the pre-processed data reflect a world that incorporates this fairness. However, we observe differences between fairadapt and warping as for warping we see a larger increase in performance while for fairadapt, the performance is higher from the start (see varying y-axis values). Yet, we have to be cautious in over-interpreting these differences in the AUC as we are more interested in the general trend towards increased performance. 
 Similarly, the AUC on the real world data is consistently higher than on pre-processed data. 
 Note that this higher performance must not be misinterpreted as favoring real-world data for evaluation; on the contrary: while the fully constrained real world classifier is now fair wrt. DP, it still carries bias from the unfair representation of the data it was trained on. 
 Instead, in order to obtain a model that incorporates a fair representation of the data, we need to train and test models on the pre-processed data and only in this way we can meaningfully evaluate performance. 
 
\begin{figure}[tbp]
    \centering
    \begin{subfigure}[b]{0.23\textwidth}
        \includegraphics[width=\linewidth]{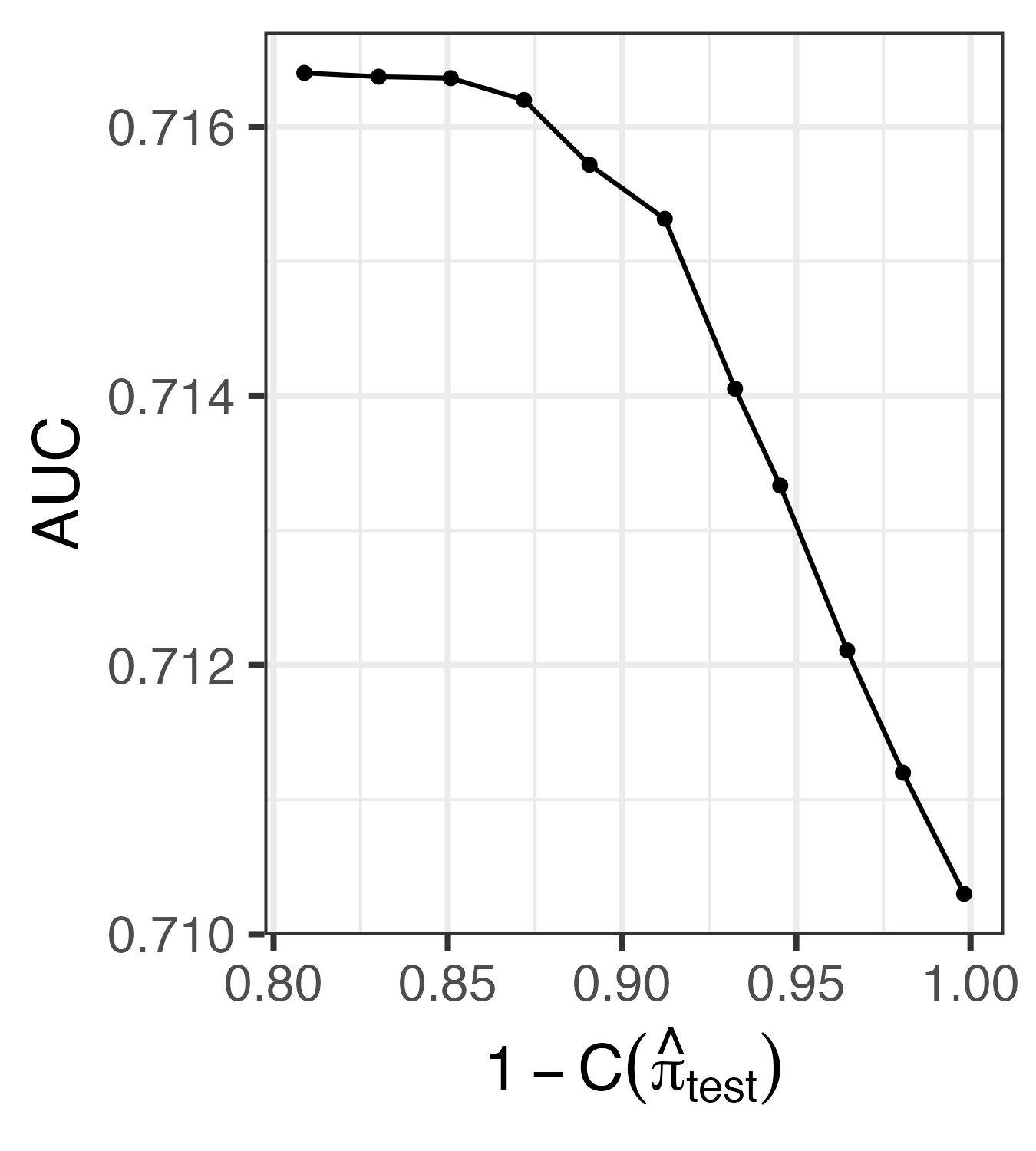}
        \caption{Real $\Dtest$} \label{fig:hmda_app_real}
    \end{subfigure}
    \hskip 1cm
    \begin{subfigure}[b]{0.23\textwidth}
        \includegraphics[width=\linewidth]{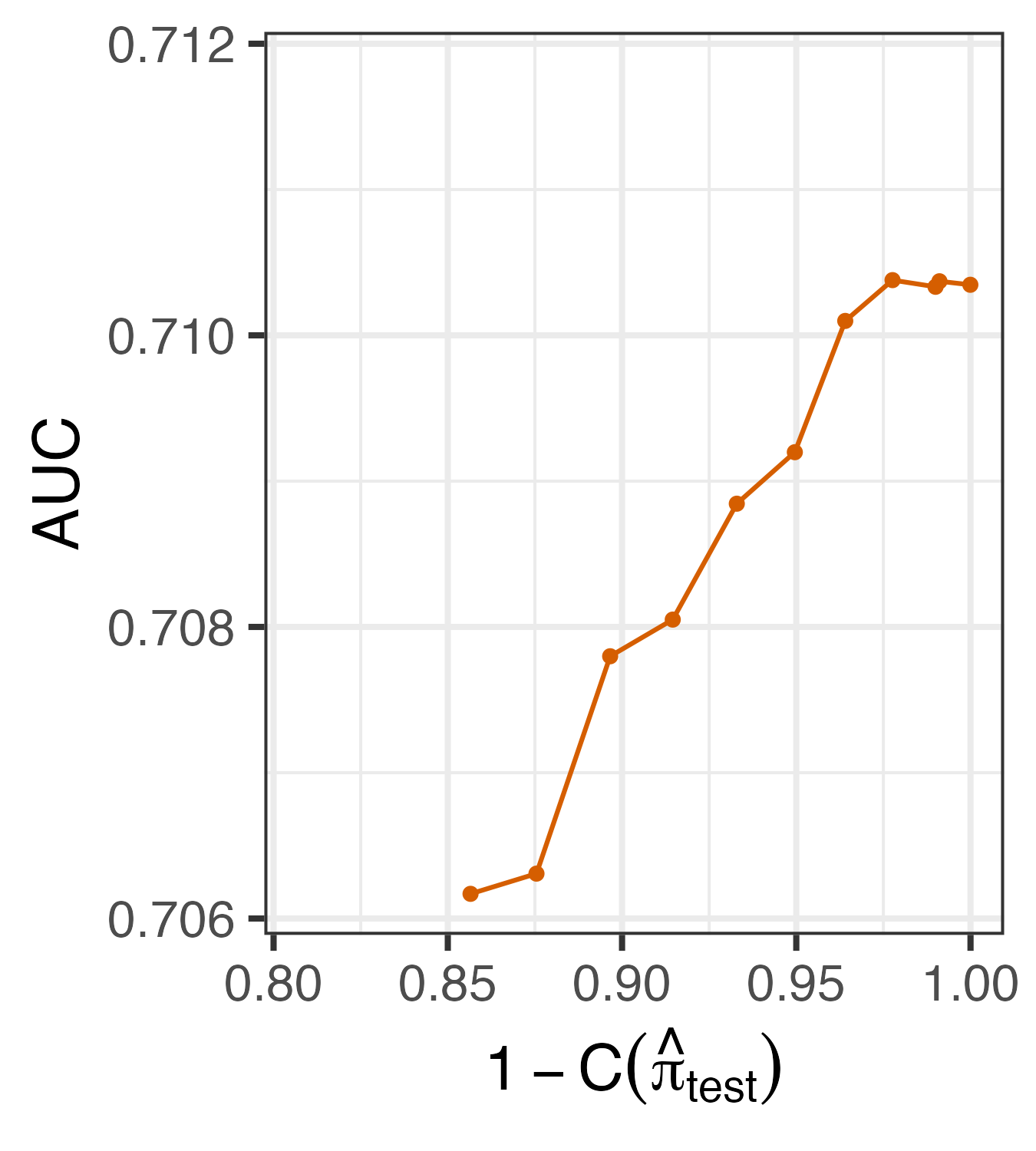}
        \caption{Adapted $\Dtest^\text{a}$} \label{fig:hmda_app_adapt}
    \end{subfigure}
    \hskip 1cm
    \begin{subfigure}[b]{0.23\textwidth}
        \includegraphics[width=\linewidth]{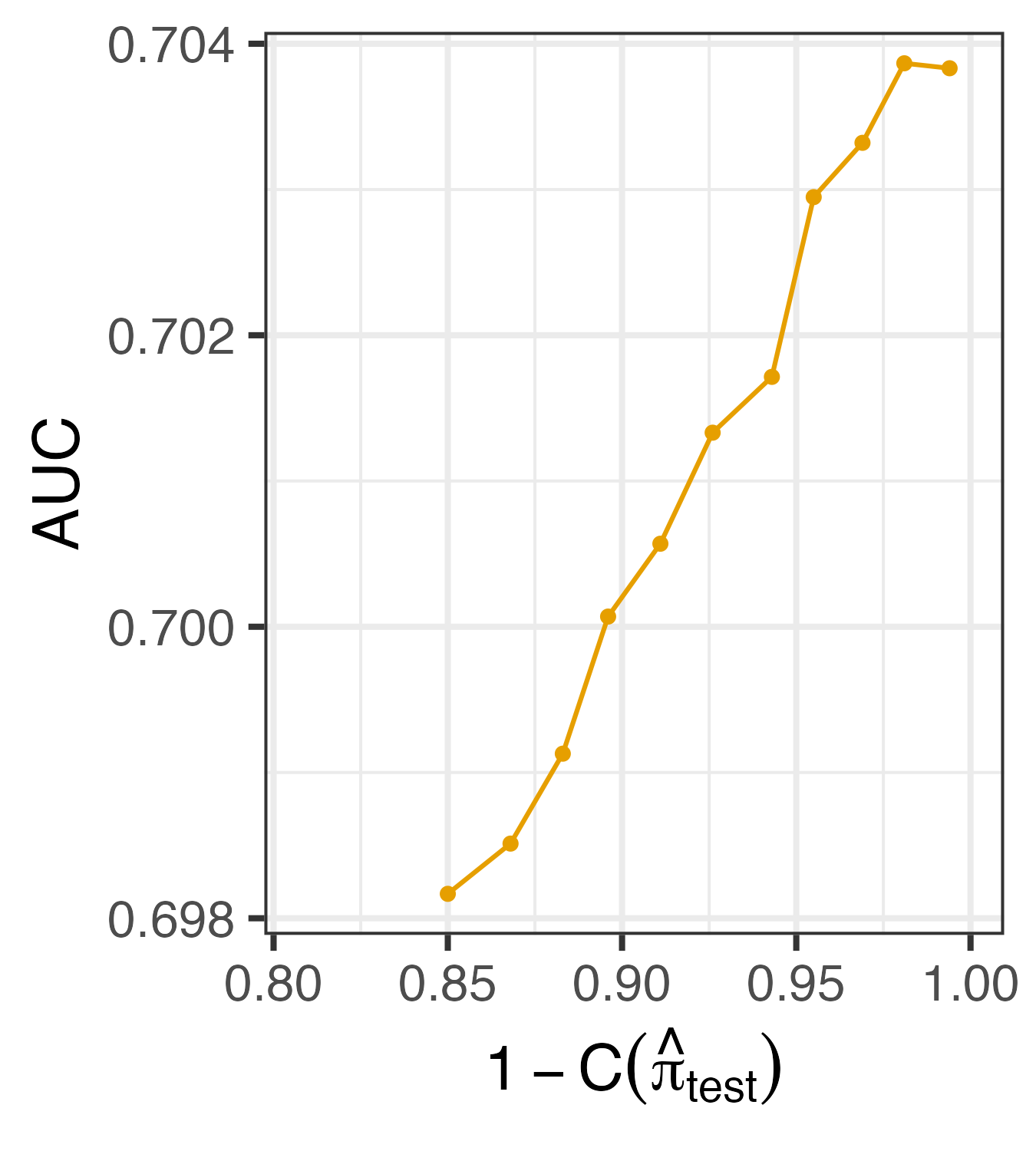}
        \caption{Warped $\Dtest^\text{w}$} \label{fig:hmda_app_warped}
    \end{subfigure}
    \caption{Fairness-performance curves of predictors trained on HMDA $\Dtrain$}
    \label{fig:comparison_tests}
\end{figure}

\subsection{Resolving the Trade-off Between Fairness Metrics} 
We observe that the pre-processing methods are also able to overcome the trade-off between several fairness metrics for the HMDA data. We proceed in the same manner as in Section~\ref{sec:fairness_fairness} by directly training and evaluating models on the pre-processed data.
The results are displayed in \autoref{tab:imp_res2}. For the real HMDA data, the fairness metrics are not simultaneously satisfied, with fairness differences up to $20\%$. 
However, for the adapted and warped data, all fairness metrics are approximately fulfilled. Regarding performance, we observe that models trained and evaluated on real, adapted, and warped data all achieve similarly high AUCs. 

\section{Summary \& Discussion}
By employing pre-processing methods that effectively approximate the FiND world, practitioners can overcome both the trade-off between different fairness metrics and the fairness-accuracy trade-off.
We have grounded this theoretically by deriving how the FiND world naturally incorporates several notions of fairness both on the individual and on the group level, including those subject to the impossibility theorem. 
This also opened up the possibility of solving the second trade-off. When learning a FiND world representation of the data via pre-processing methods, we can focus on achieving high predictive performance while being fair at the same time. 
In empirical studies on simulated and real datasets, we conclude that both the fairadapt and warping methods represent promising techniques to achieve this goal. 

However, for the purpose of our study, we considered all dependencies between the PA and the target to be unjust -- and thus eliminated all corresponding causal paths. This varies from several other proposals in fairML literature that only eliminate some path-specific effects from the PA on the target as they consider some features dependent on the PA to be resolving variables \citep{kilbertus_avoiding_2018,chiappa_pathwise_cf_2019, plecko_fair_2020}. \citet{bothmann_what_2024} also consider an alternative definition of the FiND world in which such paths can be deemed fair and need not be removed. 
Future work could investigate how to extend our findings on such alternative FiND worlds, e.g., by using the work of
\citet{plecko_fair_2020}, who propose conditional fairness metrics in such cases of resolving variables. Yet, the question of which variables can be considered resolving and non-resolving also depends on the specific use case and is not easily answered.  

The assumed knowledge of the causal graph is a necessary requirement for causal pre-processing methods, otherwise it is limited to purely statistical relations. Learning the true causal graph from data is beyond the scope of this work and forms a separate research branch known as causal discovery or causal structure learning. For a recent overview of the field, we refer the reader to \citet{squires_CSL_review_22} and \citet{kitson_BN_survey_23}.
Nevertheless, we highlight the limitation of a correctly specified DAG. In the presence of latent confounding or selection bias, the resulting independence model can no longer be expressed by a DAG over the observed variables only. Besides, the misspecification of the DAG can also degrade the effectiveness of the pre-processing methods in eliminating all causal effects from the PA on $Y$.
In the HMDA dataset, Gender can also be considered an additional PA, which we did not consider in our study.
Modeling multiple PAs opens up the topic of intersectionality (see, e.g.,~\cite{morina_auditing_2020, buolamwini_gender_2018, kearns_prev_2018}), which imposes another major challenge in fairML.

Since the FiND world represents a causal framework, we have focused on causal pre-processing methods. However, future work could include evaluating non-causal pre-processing methods in their ability to approximate the FiND world. Moreover, our study considered relatively simple DAGs. To further validate our results, applications involving structurally more complex DAGs with a larger number of variables could be explored. While the shown theory is independent from the complexity of the DAGs, it would be interesting to analyze how well the transformation methods scale. Additionally, analyzing further real-world datasets would help to further demonstrate the practical applicability and generalizability of our approach.

\section{Conclusion \& Outlook} \label{conclusion}
Our results offer practical solutions for fairness-aware machine learning, where one does not have to choose between fairness and high predictive performance anymore, nor between several classical fairness metrics. Instead, one can shift the modeling and evaluation process to an approximated FiND world, using appropriate pre-processing methods such as the fairadapt method or warping. To this end, we provide practitioners with an evaluation metric that indicates whether the applied pre-processing method has successfully approximated the FiND world. By directly training and testing models on data that approximates this FiND world, we overcome the need to explicitly enforce certain fairness metrics, as the approximation of the FiND world already incorporates a holistic assurance of these fairness metrics. Rather, we can now focus on ``just'' achieving high predictive performance.
We have shown in a simulation study that the two covered pre-processing methods are able to approximate the FiND world and showcased an application with real-world mortgage data.

Future work could investigate if other pre-processing methods could be used for approximating the FiND world and compare different pre-processing methods on diverse simulated and real-world data sets. Worthwhile directions are further the incorporation of multiple PAs, aiming at intersectionality, and an investigation of the sensitivity of the methods with respect to DAGs (partially) learned from data.

\section*{Ethical Considerations}
 Ethical and responsible application requires ongoing critical assessment of both the tools and the values guiding their use. Our methods are not a one-size-fits-all solution. Practitioners must carefully evaluate the specific use case and recognize that the assumptions underlying our methods may not apply universally. Users are also required to reflect on and consider their own normative positions, as these perspectives influence the interpretation and application of our methods.

\section*{Adverse Impact}
We emphasize the importance of careful thought when applying our methods. For practitioners, it is required to correctly apply the pre-processing methods and always consider the specific assumptions for the application at hand, in particular the causal graph. Mindless use of the methods or overtrusting their results might even have adverse impacts.

\bibliographystyle{ACM-Reference-Format}
\bibliography{tradeoff_paper.bib}
\newpage

\appendix

\section{Simulation Study} 

\subsection{Simulation Setup}
\label{app_sim_setup}
We simulate the causal relationships of the fictitious credit application example depicted in \autoref{fig:DAG}, using the R package \texttt{simcausal}~\cite{simcausal_2017}. 
 In the FiND world, we eliminate the PA's effect by setting amount $X_A$, debt $X_D$, and the target $Y$ of $A=a$ to their corresponding values among the $A=a'$ distributions $X_A^*$, $X_D^*$ and $Y^*$: 
\begin{align*}
    \textbf{Real world:} \\
    A &\sim \text{B}(\pi_A) \\
    X_C &\sim \text{Ga}(\alpha_C, \beta_C) \\
    X_A|X_C, A &\sim \text{Ga}(\alpha_A(X_C, A), \beta_A(X_C, A)) \\
    X_D|X_C, A &\sim \text{B}(\pi_D(X_C, A)) \\
    Y|X_A, X_D, X_C, A &\sim \text{B}(\pi_Y(X_A, X_D, X_C, A)) \\
    \\
    \textbf{FiND world:} \\
    A &\sim \text{B}(\pi_A) \\
    X_C &\sim \text{Ga}(\alpha_C, \beta_C) \\
    X^*_A|X_C &\sim \text{Ga}(\alpha_{Am}(X_C, a'), \beta_A(X_C, a')) \\
    X_D^*|X_C &\sim \text{B}(\pi_D(X_C, a')) \\
    Y^*|X_A^*, X_D^*, X_C &\sim \text{B}(\pi_Y(X_A^*, X_D^*, X_C, a'))
\end{align*}
In both worlds, the PA $A$ is generated by a Bernoulli distribution with success probability $\pi_A = 0.5$, while the confounder is Gamma distributed with $\alpha_C = 3.26$ and $\beta_C= 10.91$. For $\alpha_A$ and $\beta_A$, we take linear combinations of the features in combination with a log link, and for $\pi_D$ and $\pi_Y$ a logit link. We simulate datasets of size $N=10,000$ for each world, where we use the same seed for both worlds to assure comparability and perform 25 iterations. 

\subsection{Approximating the FiND world}
\label{app:approx_sim}

\begin{figure}[H]
    \centering
    \includegraphics[width=0.8\textwidth]{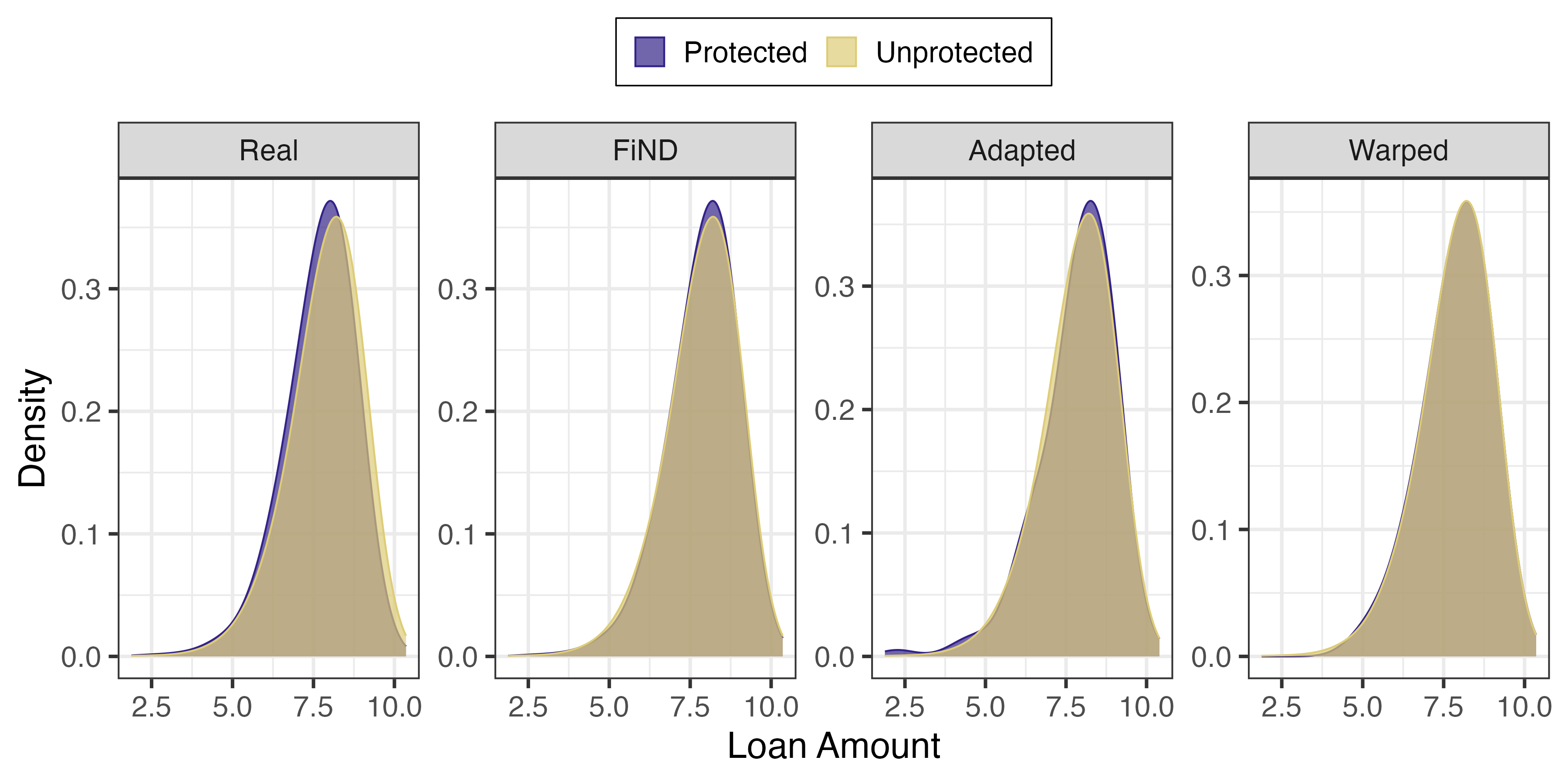}
    \caption{Distribution of $X_A$ in simulated real, FiND, adapted and warped world per protected $a$ and unprotected group $a'$}
    \label{fig:loan_amount_dist_sim}
\end{figure}

\begin{table}[H]
\centering
\caption{Distribution of $Y$ and $X_D$ in simulated real, FiND, adapted and warped world per protected $a$ and unprotected group $a'$} \label{tab:ci_sim}
    \begin{tabular}{lcccc}
        \toprule
        & \textbf{Real} & \textbf{FiND} & \textbf{Adapted} & \textbf{Warped} \\
        \midrule
        \multirow{2}{*}{$P(Y=1)$}      
            & $a = 0.755$   & $a = 0.569$   & $a = 0.585$ & $a = 0.569$ \\
            & $a' = 0.569$  & $a' = 0.569$  & $a' = 0.569$ & $a' = 0.569$ \\
        \midrule
        \multirow{2}{*}{$P(X_D=1)$}    
            & $a = 0.732$   & $a = 0.928$   & $a = 0.922$  & $a = 0.927$\\
            & $a' = 0.927$  & $a' = 0.927$  & $a' = 0.927$  & $a' = 0.927$\\
        \bottomrule
    \end{tabular}
\end{table}

\begin{table}[h]
\centering
\caption{Fairness and performance metrics in real, FiND, adapted, and warped world for the simulation study in Section~\ref{sec:sim_fairness_performance} alongside their $95\%$~ confidence intervals. All predictors are trained and evaluated using data from the same world, e.g., trained and evaluated on real world, trained and evaluated on FiND world, etc.}
\begin{tabular}{lccccc}
\toprule
\multirow{2}{*}{World} & \multicolumn{4}{c}{Fairness} & \multicolumn{1}{c}{Performance} \\
& DP & FPR & FNR & PPV & AUC \\
\midrule
Real   & 0.825$_{[0.792, 0.862]}$ & 0.782$_{[0.750, 0.836]}$ & 0.954$_{[0.926, 0.983]}$ & 0.986$_{[0.955, 0.998]}$ & 0.887$_{[0.895, 0.899]}$ \\
FiND   & 0.987$_{[0.964, 0.999]}$ & 0.989$_{[0.963, 1.000]}$ & 0.991$_{[0.976, 0.999]}$ & 0.988$_{[0.957, 0.999]}$ & 0.897$_{[0.895, 0.899]}$ \\
Adapted & 0.982$_{[0.959, 0.997]}$ & 0.972$_{[0.942, 0.995]}$ & 0.984$_{[0.954, 0.997]}$ & 0.975$_{[0.952, 0.999]}$ & 0.886$_{[0.883, 0.889]}$ \\
Warped  & 0.982$_{[0.959, 0.996]}$ & 0.964$_{[0.938, 0.992]}$ & 0.974$_{[0.943, 0.998]}$ & 0.971$_{[0.943, 0.996]}$ & 0.893$_{[0.888, 0.899]}$ \\
\bottomrule
\end{tabular}
\label{tab:imp_res_ci}
\end{table}

\section{HMDA Experiment} \label{app_ex_setup}

\subsection{Data Setup}
We encode and filter the 2022 Home Mortgage Disclosure Act (HMDA) data of the state Wisconsin in the following way\footnote{A detailed description of all variables is provided here: \url{https://ffiec.cfpb.gov/documentation/publications/loan-level-datasets/lar-data-fields}}: 

\begin{itemize}
\item $Y$: Binary target indicating loan approved (1) or not approved (0). The original variable ``action taken'' has eight categories and encodes the status of the loan.
\item $A$: Binary PA race with levels $a$ Black applicant or $a'$ non-Hispanic White applicant. 
\item $X_A$: Numerical variable of the amount of the covered loan, log-transformed. 
\item $X_P$: Binary variable indicating the purpose of the loan, (1) home purchase or not (0). The original variable has four categories. 
\item $X_D$: The debt to income ratio, with binary category (1) high ratio or not (0). 
\item $X_C$: The joint confounders age and gender. Binary age indicates (1) age above 62 or not (0). Binary gender indicates (1) female or not (0). (Note that gender is assumed to be binary purely for simplicity reasons and does not reflect the authors' personal view.)
\end{itemize}

\subsection{Approximating the FiND world}
\label{app:approx_HMDA}

\begin{figure}[H]
    \centering
    \includegraphics[width=0.6\textwidth]{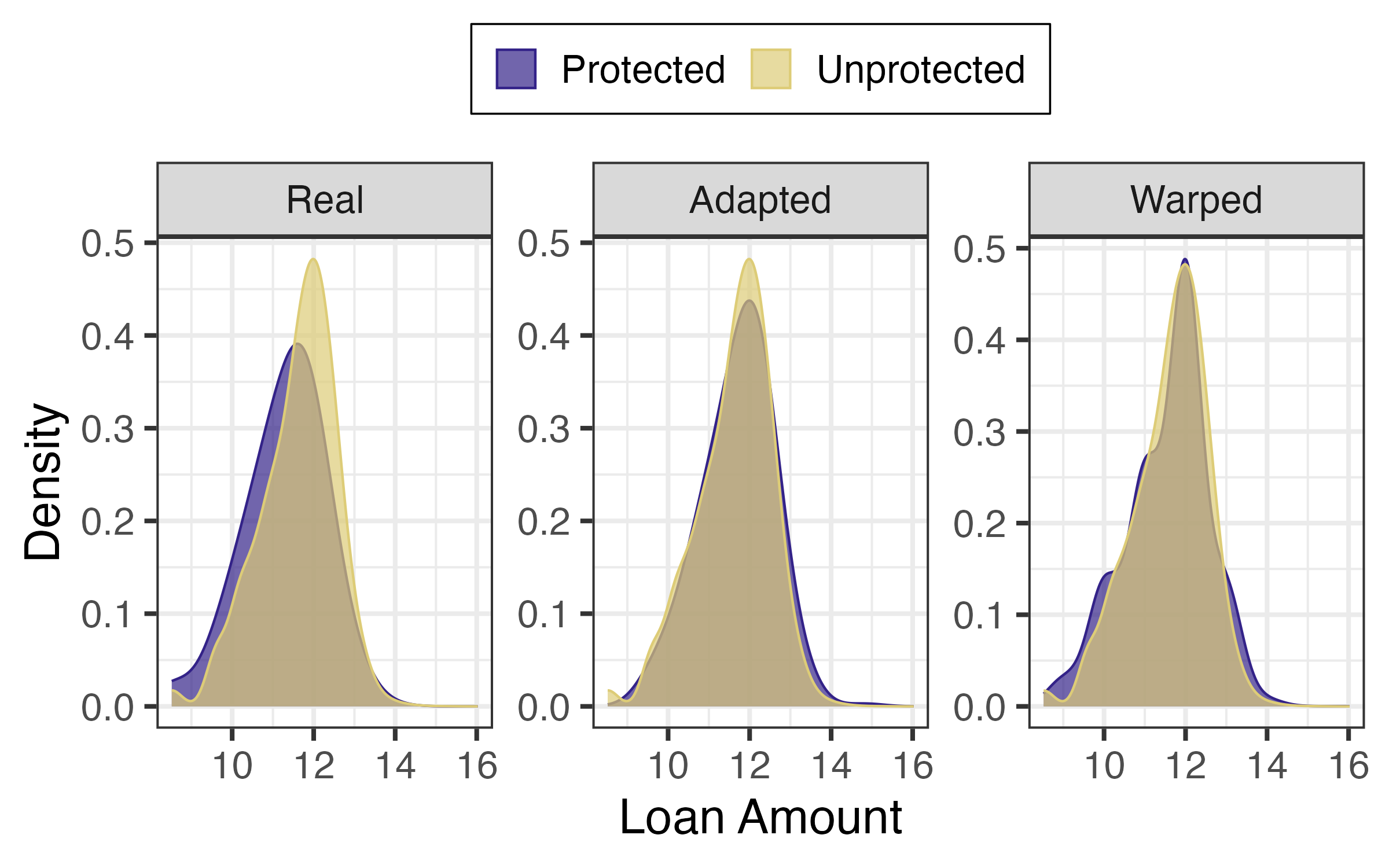}
    \caption{Distribution of $X_A$ on real-world, adapted and warped HMDA data per protected $a$ and unprotected group $a'$}
    \label{fig:loan_amount_dist}
\end{figure}

\begin{table}[H]
    \centering
    \caption{Distribution of $Y$, $X_P$ and $X_D$ on real-world, adapted and warped HMDA data per protected $a$ and unprotected group $a'$}
    \begin{tabular}{lccc}
        \toprule
        & \textbf{Real} & \textbf{Adapted} & \textbf{Warped} \\
        \midrule
        \multirow{2}{*}{$P(Y=1)$}      
            & $a = 0.486$   & $a = 0.677$   & $a = 0.679$  \\
            & $a' = 0.679$  & $a' = 0.679$  & $a' = 0.679$  \\
        \midrule
        \multirow{2}{*}{$P(X_P=1)$}      
            & $a = 0.347$   & $a = 0.398$   & $a = 0.384$  \\
            & $a' = 0.390$  & $a' = 0.390$  & $a' = 0.390$  \\
        \midrule
        \multirow{2}{*}{$P(X_D=1)$}    
            & $a = 0.702$   & $a = 0.665$   & $a = 0.629$  \\
            & $a' = 0.629$  & $a' = 0.629$  & $a' = 0.629$  \\
        \bottomrule
    \end{tabular}
\end{table}

\end{document}